\documentclass{article} 

\usepackage[preprint]{colm2026_conference}
\usepackage{microtype}
\usepackage{hyperref}
\usepackage{url}
\usepackage{booktabs}
\usepackage{graphicx}
\usepackage{amsmath, amssymb}
\usepackage{colortbl}
\usepackage[normalem]{ulem}  
\usepackage{makecell}        
\usepackage{multirow}   
\usepackage{array}      
\usepackage{caption}    
\usepackage{tcolorbox}
\usepackage{wrapfig}
\usepackage{float}

\definecolor{darkblue}{rgb}{0, 0, 0.5}
\hypersetup{colorlinks=true, citecolor=darkblue, linkcolor=darkblue, urlcolor=darkblue}

\tcbuselibrary{skins, breakable}

\newtcolorbox{cotbox}[1][]{
    breakable,             
    enhanced,
    colback=gray!5,        
    colframe=gray!40,      
    boxrule=0.5pt,         
    arc=2mm,               
    left=2mm, right=2mm, top=2mm, bottom=2mm,
    fonttitle=\bfseries\small\sffamily, 
    coltitle=black,
    title=#1               
}

\title{CoT-Core: Accelerating LLM Evaluation via CoT-Aware Coreset Selection}

\author{
  \textbf{Qihua Pan}$^1$, \textbf{Zhenheng Tang}$^2$, \textbf{Peijie Dong}$^3$, \textbf{Xiang Liu}$^3$, \\
  \textbf{Huacan Wang}$^4$, \textbf{Bo Li}$^2$, \textbf{Xiaowen Chu}$^{3, \ast}$ \\
  $^1$Nanjing University \\
  $^2$The Hong Kong University of Science and Technology \\
  $^3$The Hong Kong University of Science and Technology (Guangzhou) \\
  $^4$Independent Researcher \\
}

\begin{document}

\maketitle

\begin{abstract}
Evaluating Large Language Models (LLMs) incurs prohibitive computational overhead during continuous development processes. While coreset selection accelerates evaluation, existing methods either suffer from a severe ``cold start'' bottleneck requiring massive historical logs (e.g., Item Response Theory) or exhibit a surface lexical bias that misses the underlying reasoning manifold of tasks. We propose CoT-Core, a novel training-free core question selection framework. Recognizing that lexically disparate questions can share equivalent underlying logic, CoT-Core prompts LLMs to unroll zero-shot Chain-of-Thought (CoT) reasoning trajectories. Projecting these paths into a latent space effectively clusters questions by intrinsic logical equivalence rather than superficial text similarity. Extensive experiments on GSM8K, MMLU, MMLU-Pro, and GPQA demonstrate that CoT-Core drastically reduces evaluation costs while maintaining high-fidelity score estimation, and delineate the boundary conditions of reasoning-aware pruning, revealing that its efficacy is intrinsically gated by task complexity.
\end{abstract}


\section{Introduction}
\label{sec:intro}
The rapid advancement of Large Language Models (LLMs) \citep{achiam2023gpt, touvron2023llama2} has necessitated the development of comprehensive evaluation frameworks and benchmarks, such as MMLU \citep{hendrycks2020measuring}, GSM8K \citep{cobbe2021training}, and HELM \citep{liang2022holistic}. However, evaluating modern LLMs across tens of thousands of test cases introduces prohibitive computational and financial overhead \citep{srivastava2023beyond}. Furthermore, evaluation is rarely a one-time event; during the continuous pre-training, alignment, and fine-tuning phases, frequent and repeated evaluations across numerous model checkpoints are strictly required to monitor progress and steer model development \citep{biderman2023pythia, groeneveld2024olmo}, severely exacerbating the consumption of computational resources.

Recent literature highlights that this exhaustive evaluation paradigm is highly inefficient due to inherent dataset redundancy \citep{wang2025rethinking, zhang2025redundancy}. This creates an urgent need for \textit{coresets}---minimal subsets of data that accurately reconstruct full-dataset evaluation scores. While various sampling heuristics exist, the most prominent and systematic approaches for constructing evaluation coresets generally align with two main paradigms. Psychometric methods, which score questions based on historical model correctness using Item Response Theory (IRT) \citep{polo2024tinybenchmarks}, are fundamentally bottlenecked by their heavy reliance on massive, pre-existing evaluation logs. Consequently, they face a severe ``cold start'' problem: for newly constructed benchmarks or private evaluation datasets, the extensive historical response data required to calibrate item parameters is simply unavailable. This strict prerequisite renders history-dependent methods completely inapplicable to emerging evaluation scenarios where ground-truth correctness records across a wide range of models have not yet been accumulated. On the other hand, geometric methods, such as the widely adopted $k$-center greedy algorithm \citep{sener2017active}, typically operate on shallow textual embeddings of the questions. While computationally efficient, these dense encoders exhibit a strong bias toward surface lexical overlap \citep{sinha2021unnatural, thakur2021beir}, fundamentally failing to capture the latent reasoning structures underlying the tasks.

\begin{wrapfigure}{R}{0.48\columnwidth}
    \centering
    \vspace{-10pt} 
    \includegraphics[width=\linewidth]{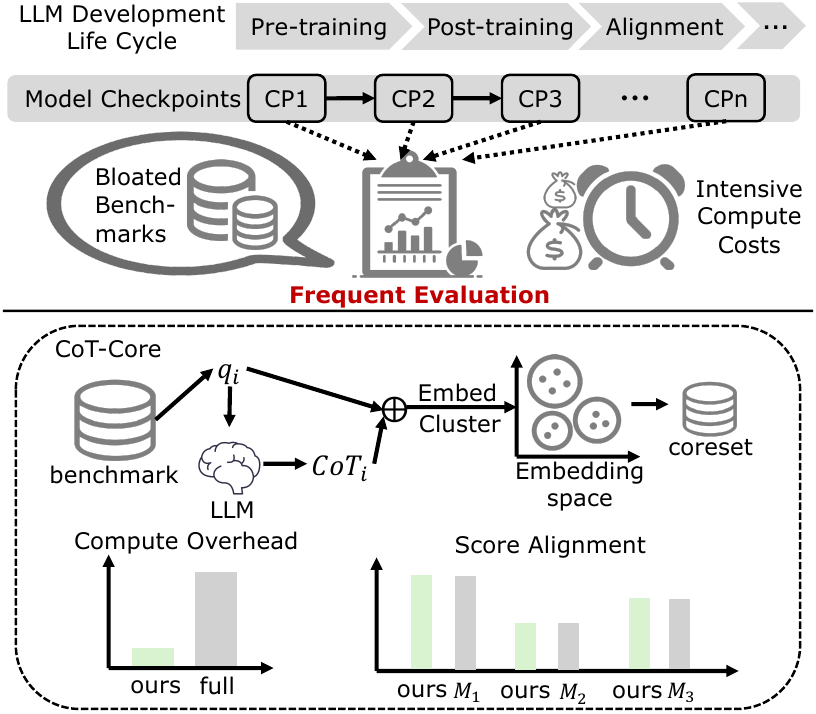}
    \caption{The motivation and overview of CoT-Core.}
    \label{fig:motivation_overview}
    \vspace{-10pt} 
\end{wrapfigure}

To overcome the limitations of history-dependent metrics and shallow lexical features, we propose CoT-Core, a novel training-free core question selection framework. Our approach is motivated by a critical observation: lexically disparate questions---which appear entirely unrelated in standard text embedding spaces---can share equivalent underlying logic or reasoning structures. To capture this underlying reasoning manifold, CoT-Core first prompts LLMs to unroll explicit Chain-of-Thought (CoT) reasoning trajectories. By directly projecting these unrolled reasoning paths into a dense embedding space, CoT-Core effectively overcomes lexical divergence, achieving highly precise and cognitive-aligned question clustering. Building upon this logically structured latent space, we apply standard clustering algorithms to extract a minimal, highly representative core subset. This selected coreset serves as a high-fidelity proxy for the full benchmark, drastically mitigating the computational and financial overhead of continuous model evaluation while maintaining accurate performance estimation.

Our main contributions are summarized as follows: 1) We propose \textbf{CoT-Core}, a novel training-free coreset selection framework that circumvents surface-level \textit{lexical traps}. By projecting unrolled reasoning trajectories into a latent space, it clusters questions based on their underlying \textit{reasoning manifold} rather than superficial text features. 2) Methodologically, we decouple spatial coreset selection from downstream proxy estimation. CoT-Core employs centroid-based geometric sampling to extract representative instances, naturally supporting Standard Aggregation (SA) for a strictly training-free pipeline, while remaining seamlessly compatible with advanced probabilistic estimators (e.g., GP-IRT). 3) Extensive experiments demonstrate that CoT-Core drastically reduces evaluation overhead while maintaining high-fidelity score estimation. Furthermore, we delineate the boundary conditions of reasoning-aware pruning, revealing that its efficacy is intrinsically tied to task complexity.

\section{Related Work}
\label{sec:related}

\paragraph{LLM Benchmark Evaluation and Redundancy}
As the demand for rigorous assessment grows, evaluating models across massive suites has become computationally bottlenecked. Studies such as EssenceBench \citep{wang2025rethinking} have systematically quantified sample redundancy within prominent evaluation suites, revealing significant overlap at both the textual level and the ranking level. Similarly, \citet{zhang2025redundancy} demonstrated that repeated testing of similar capabilities yields diminishing returns. Identifying a representative coreset is therefore an urgent necessity for sustainable AI development.

\paragraph{Geometric and Training-free Methods.} 
As a foundational approach to eliminate redundancy without historical reliance, geometric algorithms like the $k$-center greedy method \citep{sener2017active} prioritize maximizing spatial manifold coverage. However, these methods traditionally construct feature spaces using shallow embeddings of \textit{raw problem text}. Consequently, they are susceptible to surface-level lexical traps: while two problems might share identical vocabulary, they often necessitate completely different reasoning trajectories, leading to suboptimal pruning that misses structural diversity.

\paragraph{Performance-Based and Psychometric Methods.} 
Alternatively, another research branch attempts to capture latent item difficulty through historical model performance. Classic paradigms filter instances based on training dynamics \citep{swayamdipta2020dataset}, while recent advancements like TinyBenchmarks \citep{polo2024tinybenchmarks} and ATLAS \citep{li2025adaptive} leverage Item Response Theory (IRT) to estimate parameters from large-scale response matrices. While effective at capturing cognitive depth, these \textit{history-dependent} methods suffer from a severe ``cold start'' problem. Their reliance on extensive historical records renders them inapplicable to emerging or newly constructed evaluation suites where such data is entirely absent.

\paragraph{CoT-Aware Representations for Coreset Selection}
Chain-of-Thought (CoT) prompting \citep{wei2022chain} transforms a black-box mapping into a transparent sequence of cognitive operations. While previous works (e.g., Auto-CoT \citep{zhang2022automatic}) combine embeddings and clustering, they predominantly operate on the embeddings of the \textit{raw questions}, conflating superficial lexical overlap with true structural similarity. Recent studies demonstrate that LLM reasoning follows globally coherent abstract paths, which are far better captured by embedding the reasoning process itself \citep{yu2025explainable}. 

In this work, we propose a paradigm shift by utilizing CoT trajectories and embeddings as a unified feature space. Instead of surface text, CoT-Core extracts the \textbf{latent logical structure} of the reasoning path, abstracting away specific numerical values and concrete entities. By clustering these \textbf{isomorphic reasoning structures}, we group questions necessitating identical cognitive meta-operations. Crucially, because this representation intrinsically captures cognitive difficulty and latent logic, our clustering approach remains entirely \textit{training-free}.

\section{Methodology}
\label{sec:method}

In this section, we present the technical details of \textbf{CoT-Core}. We first formally define the evaluation coreset problem, and then detail the three primary stages of our pipeline: CoT Trajectory Generation, Contextualized Trajectory Embedding, and Isomorphic Clustering and Selection.

\subsection{Problem Formulation}
\label{subsec:problem}
Let $\mathcal{D} = \{(q_i, a_i)\}_{i=1}^N$ denote a comprehensive LLM evaluation benchmark comprising $N$ test instances, where $q_i$ is the question prompt and $a_i$ is the reference answer. For a given large language model $\mathcal{M}$, the standard evaluation process computes its true overall performance by averaging the exact match scores across the entire dataset:
\begin{equation}
    S_{\mathcal{D}}(\mathcal{M}) = \frac{1}{N} \sum_{i=1}^N m(\mathcal{M}(q_i), a_i)
\end{equation}
where $\mathcal{M}(q_i)$ represents the model's generated output and $m(\cdot, \cdot)$ is a specific evaluation metric function (e.g., exact match accuracy).

The overarching objective of coreset evaluation intrinsically consists of two orthogonal stages: \textbf{Subset Selection} and \textbf{Performance Estimation}. 

First, the selection stage identifies a minimal representative subset $\mathcal{S} \subset \mathcal{D}$, strictly bounded by a budget constraint $|\mathcal{S}| = B \ll N$. Second, the estimation stage employs a proxy scoring function, denoted as $f_{est}$, to predict the model's full-dataset capability based solely on its responses to the constrained subset $\mathcal{S}$. 

The ultimate goal is to formulate a subset $\mathcal{S}$ and an estimator $f_{est}$ such that the predicted proxy score closely approximates the true full-dataset performance for any unseen model $\mathcal{M}$:
\begin{equation}
    \hat{S}_{\mathcal{D}}(\mathcal{M}) = f_{est}(\mathcal{M}, \mathcal{S}) \approx S_{\mathcal{D}}(\mathcal{M})
\end{equation}
Unlike history-dependent baselines that tightly couple their selection algorithms with specific parameterized estimators, CoT-Core operates fundamentally as an \textit{estimator-agnostic} framework. Our primary objective is to deliver a highly informative subset $\mathcal{S}$ in a strictly training-free setting. The resulting coreset is inherently flexible: it naturally supports Standard Aggregation (SA) as $f_{est}$ for a deterministic baseline, while remaining fully compatible with advanced probabilistic estimators if historical data is accessible.

\subsection{CoT-Core Framework Overview}
\label{subsec:overview}

\begin{figure*}[t]
    \centering
    \includegraphics[width=0.9\textwidth]{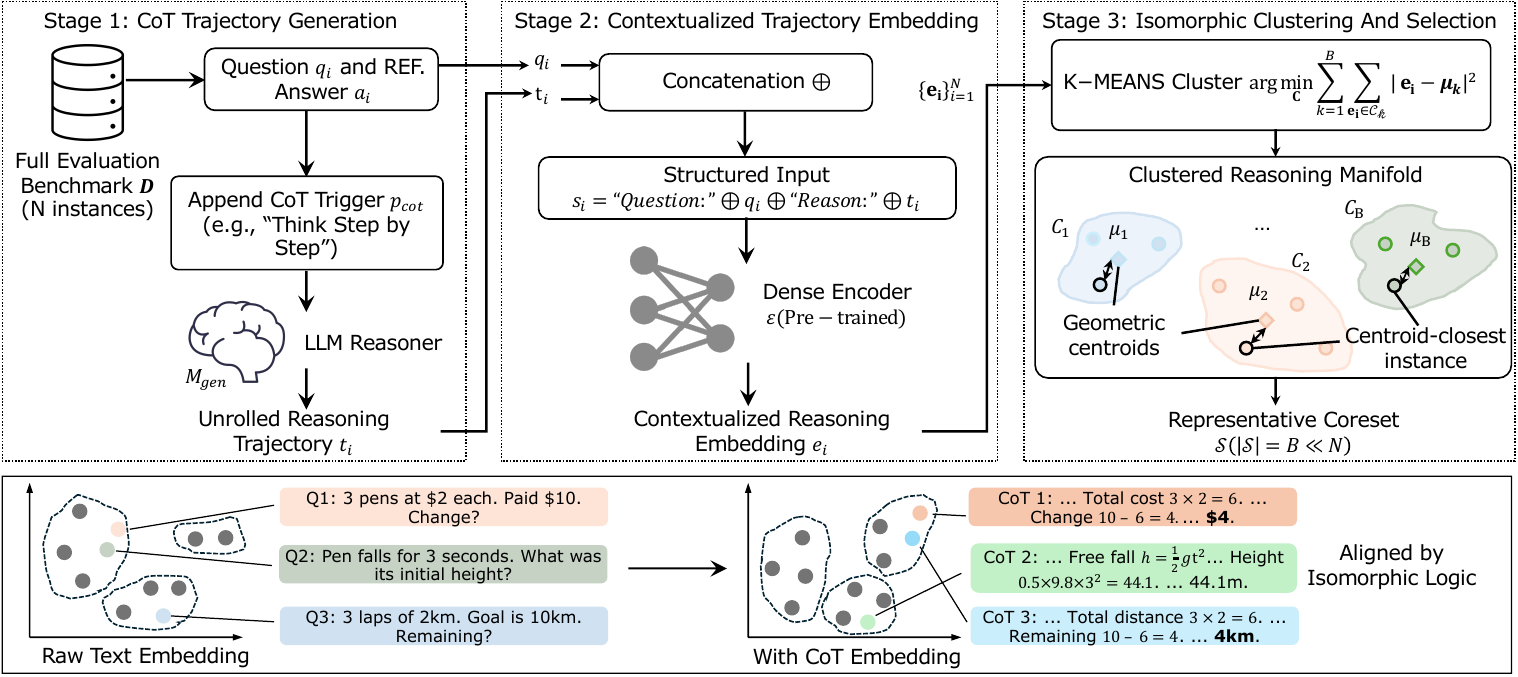}
    \caption{\textbf{Overview of CoT-Core.} Our framework unrolls LLM reasoning trajectories, embeds them, and applies isomorphic clustering to select coresets. In contrast to raw text embeddings which fall into lexical traps (falsely grouping Q1 \& Q2), CoT-Core effectively aligns lexically disparate problems that share an identical latent logical structure (Q1 \& Q3).}
    \label{fig:framework}
\end{figure*}

The severe absence of historical data in real-world scenarios forces coreset construction into a strict training-free setting. Under this severe constraint, conventional text embeddings fundamentally fail to capture the underlying logic. Compelled by this limitation, CoT-Core leverages unrolled \textit{reasoning trajectories} as a necessary proxy to measure true semantic distance.

As illustrated in Figure \ref{fig:framework}, our framework systematically maps the reasoning manifold of evaluation datasets through a straightforward, three-stage pipeline: (1) eliciting zero-shot reasoning trajectories via a generator LLM (Section \ref{subsec:stage1}), (2) projecting these contextualized paths into a dense latent space (Section \ref{subsec:stage2}), and (3) applying training-free geometric clustering to extract the final representative coreset (Section \ref{subsec:stage3}).

\subsection{Stage 1: CoT Trajectory Generation}
\label{subsec:stage1}
As highlighted in our motivation, embedding the raw question text $q_i$ captures only superficial lexical features (e.g., vocabulary overlap), fundamentally failing to capture its underlying logic. To address this, our first stage shifts the representation space from the problem definition to the problem resolution path. 

For each instance in the dataset $\mathcal{D}$, we utilize an instruction-tuned LLM as the generator, denoted as $\mathcal{M}_{gen}$. We append a standard Chain-of-Thought trigger (e.g., \textit{``Let's think step by step''}) to the question $q_i$, compelling the model to articulate its complete reasoning process. The generated trajectory is defined as:
\begin{equation}
    t_i = \mathcal{M}_{gen}(q_i \oplus p_{cot})
\end{equation}
where $\oplus$ denotes string concatenation and $p_{cot}$ is the CoT prompt. Crucially, we retain this trajectory $t_i$ regardless of whether the final derived answer is correct. The unrolled path itself---even if ultimately flawed---explicitly exposes the underlying logic and step-by-step reasoning structure required to navigate the problem space.

\subsection{Stage 2: Contextualized Trajectory Embedding}
\label{subsec:stage2}
Instead of relying on history-dependent performance metrics, we directly leverage the structural density of the generated reasoning paths. Crucially, to ensure that the unrolled logic remains grounded in its original premise, we do not embed the trajectory in isolation. For each instance, we concatenate the raw question $q_i$ and the generated trajectory $t_i$ into a structured input sequence:
\begin{equation}
    s_i = \text{``Question: ''} \oplus q_i \oplus \text{`` Reason: ''} \oplus t_i
\end{equation}
where $\oplus$ denotes string concatenation. This contextualized sequence $s_i$ is then mapped into a continuous, high-dimensional space using a pre-trained dense retrieval model $\mathcal{E}$, yielding the reasoning feature matrix $\mathbf{E} \in \mathbb{R}^{N \times d}$:
\begin{equation}
    \mathbf{e}_i = \mathcal{E}(s_i)
\end{equation}
By embedding the logically unrolled CoT explicitly grounded in its problem context, $\mathbf{e}_i$ naturally captures the underlying reasoning manifold of the task. This effectively overcomes lexical divergence between lexically disparate but logically isomorphic problems.

\subsection{Stage 3: Isomorphic Clustering and Selection}
\label{subsec:stage3}
Because the embedding space now inherently encodes the underlying logic and structural equivalence, we apply standard $k$-means clustering to partition the $N$ trajectory embeddings into $B$ clusters, aligning with the coreset budget constraint. For each cluster, we select the problem instance whose reasoning embedding is closest to the geometric centroid. This training-free sampling strategy maximizes coverage of the reasoning manifold and systematically eliminates dataset redundancy without requiring prior evaluation data or trainable parameters.

Crucially, CoT-Core decouples the coreset selection $\mathcal{S}$ from the downstream proxy scoring function $f_{est}$. By default, it naturally accommodates Standard Aggregation (SA)---a deterministic baseline that applies selection-derived instance weights or reduces to an unweighted empirical mean. Furthermore, this formulation remains seamlessly compatible with advanced probabilistic models (e.g., GP-IRT) for parameterized estimation when historical response data is available.

\section{Experiments}
\label{sec:experiments}

\subsection{Experimental Setup}
\label{subsec:setup}

\textbf{Datasets \& Implementation Details.} We evaluate CoT-Core on four widely adopted reasoning benchmarks: GSM8K \citep{cobbe2021training}, GPQA (Diamond split) \citep{rein2024gpqa}, MMLU \citep{hendrycks2020measuring}, and MMLU-Pro \citep{wang2024mmlu}. For trajectory generation ($\mathcal{M}_{gen}$), we employ the open-weight Phi-4 model \citep{abdin2024phi} via the OpenCompass framework \citep{contributors2023opencompass} to ensure standardized reproducibility. The contextualized trajectories are then mapped into a continuous feature space using the BGE-M3 encoder \citep{chen2024bge} ($\mathcal{E}$).

\textbf{Baselines \& Data Splits.} We compare CoT-Core against two paradigms of coreset selection. \textit{Training-free baselines} include \textbf{Random} sampling, \textbf{Question-Emb K-Means} (clustering raw problem text embeddings), and the \textbf{$k$-Center Greedy} algorithm \citep{sener2017active}. \textit{History-dependent baselines} include \textbf{Correctness K-Means} (clustering historical binary response matrices) and \textbf{IRT} \citep{polo2024tinybenchmarks}. Evaluation logs are sourced from TinyBenchmarks \citep{polo2024tinybenchmarks} and the Open LLM Leaderboard \citep{beeching2023open}. For a rigorous evaluation of generalization, we hold out 100 disjoint models as the test set. For history-dependent methods, we fit latent parameters using a strictly disjoint training set of $N=25$ models to simulate practical low-resource settings (results for larger $N$ are deferred to Appendix \ref{sec:appendix_extended_results}).

\textbf{Evaluation Protocol.} Coresets are extracted at strict budgets of $1\%, 5\%, 10\%, 15\%$, and $20\%$ of the full dataset size. To isolate selection quality from downstream aggregation, all subsets are evaluated using two decoupled estimators: \textbf{Standard Aggregation (SA)}---a deterministic baseline that applies selection-derived instance weights or reduces to an unweighted empirical mean---and \textbf{GP-IRT} \citep{polo2024tinybenchmarks}, a parameterized historical estimator. The discrepancy between proxy estimations and true full-dataset scores is quantified via \textbf{Mean Absolute Error (MAE)} ($\downarrow$) and \textbf{Ranking Similarity (Sim)} ($\uparrow$). To rigorously mitigate initialization variance, we report the mean performance across 5 independent random seeds for all algorithmic methods, and 20 independent trials for the Random baseline.

\subsection{Main Results}
\label{subsec:results}

\begin{table*}[t]
\centering
\caption{Performance comparison at $N=25$ using the SA estimator. Best performance among training-free methods is highlighted in \textbf{bold}. History-dependent methods are shown for reference only. Note that results at the 1\% budget for GPQA are omitted, as this strictly equates to a single instance, rendering coreset evaluation statistically meaningless.}
\label{table:25sa}
\resizebox{\textwidth}{!}{
\begin{tabular}{lcccccc}
\toprule
\multirow{2}{*}{\textbf{Benchmark}} & \multirow{2}{*}{\textbf{Method}} & \multicolumn{5}{c}{\textbf{Sampling Ratio}} \\
\cmidrule(lr){3-7}
& & \makecell{1\% \\ \scriptsize MAE ($\downarrow$) / Sim ($\uparrow$)} & \makecell{5\% \\ \scriptsize MAE ($\downarrow$) / Sim ($\uparrow$)} & \makecell{10\% \\ \scriptsize MAE ($\downarrow$) / Sim ($\uparrow$)} & \makecell{15\% \\ \scriptsize MAE ($\downarrow$) / Sim ($\uparrow$)} & \makecell{20\% \\ \scriptsize MAE ($\downarrow$) / Sim ($\uparrow$)} \\
\midrule
& \multicolumn{6}{c}{\cellcolor{gray!15}\textit{Training-Free Methods (Zero-Shot)}} \\
GSM8K & Random & 0.0890 / 0.9044 & 0.0382 / \textbf{0.9747} & 0.0246 / 0.9851 & 0.0187 / \textbf{0.9897} & 0.0142 / 0.9920 \\
 & Question-Emb K-Means & 0.0875 / 0.8951 & \textbf{0.0321} / 0.9743 & 0.0241 / 0.9797 & \textbf{0.0175} / 0.9871 & \textbf{0.0133} / \textbf{0.9931} \\
 & k-Center Greedy & \textbf{0.0683} / \textbf{0.9118} & 0.0349 / 0.9738 & 0.0283 / 0.9847 & 0.0218 / \textbf{0.9897} & 0.0175 / 0.9925 \\
\rowcolor{orange!15} \cellcolor{white}  & \textbf{CoT-Feature (Ours)} & 0.0755 / 0.8884 & 0.0331 / 0.9696 & \textbf{0.0209} / \textbf{0.9853} & 0.0197 / 0.9884 & 0.0154 / 0.9918 \\
\cmidrule(lr){2-7}
& \multicolumn{6}{c}{\cellcolor{gray!15}\textit{Trainable Methods (Require Historical Data, N=25)}} \\
\rowcolor{blue!10} \cellcolor{white}  & Correctness K-Means & 0.0596 / 0.9129 & 0.0347 / 0.9700 & 0.0259 / 0.9770 & 0.0231 / 0.9838 & 0.0198 / 0.9886 \\
\rowcolor{blue!10} \cellcolor{white}  & IRT & 0.0664 / 0.9172 & 0.0308 / 0.9649 & 0.0196 / 0.9817 & 0.0175 / 0.9852 & 0.0145 / 0.9879 \\
\midrule
& \multicolumn{6}{c}{\cellcolor{gray!15}\textit{Training-Free Methods (Zero-Shot)}} \\
MMLU & Random & 0.0330 / \textbf{0.9489} & 0.0149 / 0.9880 & 0.0102 / 0.9935 & \textbf{0.0079} / 0.9957 & 0.0063 / 0.9969 \\
 & Question-Emb K-Means & \textbf{0.0298} / 0.9484 & 0.0215 / 0.9877 & 0.0154 / 0.9923 & 0.0203 / 0.9943 & 0.0252 / 0.9958 \\
 & k-Center Greedy & 0.0477 / 0.9461 & 0.0530 / 0.9867 & 0.0393 / 0.9909 & 0.0360 / 0.9942 & 0.0317 / 0.9952 \\
\rowcolor{orange!15} \cellcolor{white}  & \textbf{CoT-Feature (Ours)} & 0.0324 / 0.9353 & \textbf{0.0118} / \textbf{0.9881} & \textbf{0.0081} / \textbf{0.9941} & 0.0081 / \textbf{0.9964} & \textbf{0.0060} / \textbf{0.9971} \\
\cmidrule(lr){2-7}
& \multicolumn{6}{c}{\cellcolor{gray!15}\textit{Trainable Methods (Require Historical Data, N=25)}} \\
\rowcolor{blue!10} \cellcolor{white}  & Correctness K-Means & 0.0304 / 0.9616 & 0.0203 / 0.9754 & 0.0181 / 0.9773 & 0.0169 / 0.9781 & 0.0160 / 0.9785 \\
\rowcolor{blue!10} \cellcolor{white}  & IRT & 0.0253 / 0.9589 & 0.0112 / 0.9895 & 0.0081 / 0.9938 & 0.0068 / 0.9966 & 0.0056 / 0.9969 \\
\midrule
& \multicolumn{6}{c}{\cellcolor{gray!15}\textit{Training-Free Methods (Zero-Shot)}} \\
MMLU-Pro & Random & 0.0311 / 0.9556 & 0.0132 / \textbf{0.9899} & 0.0093 / 0.9941 & \textbf{0.0070} / 0.9956 & 0.0058 / 0.9967 \\
 & Question-Emb K-Means & 0.0300 / 0.9579 & 0.0133 / 0.9890 & 0.0111 / \textbf{0.9947} & 0.0116 / 0.9957 & 0.0100 / \textbf{0.9969} \\
 & k-Center Greedy & 0.0369 / 0.9524 & 0.0432 / 0.9857 & 0.0334 / 0.9923 & 0.0306 / 0.9936 & 0.0272 / 0.9957 \\
\rowcolor{orange!15} \cellcolor{white}  & \textbf{CoT-Feature (Ours)} & \textbf{0.0296} / \textbf{0.9682} & \textbf{0.0124} / 0.9893 & \textbf{0.0082} / 0.9941 & 0.0072 / \textbf{0.9965} & \textbf{0.0055} / \textbf{0.9969} \\
\cmidrule(lr){2-7}
& \multicolumn{6}{c}{\cellcolor{gray!15}\textit{Trainable Methods (Require Historical Data, N=25)}} \\
\rowcolor{blue!10} \cellcolor{white}  & Correctness K-Means & 0.0257 / 0.9789 & 0.0144 / 0.9902 & 0.0123 / 0.9929 & 0.0107 / 0.9936 & 0.0095 / 0.9941 \\
\rowcolor{blue!10} \cellcolor{white}  & IRT & 0.0379 / 0.9540 & 0.0151 / 0.9891 & 0.0111 / 0.9930 & 0.0092 / 0.9943 & 0.0073 / 0.9962 \\
\midrule
& \multicolumn{6}{c}{\cellcolor{gray!15}\textit{Training-Free Methods (Zero-Shot)}} \\
GPQA & Random & - / - & 0.1252 / 0.2851 & 0.0797 / 0.4480 & 0.0635 / 0.5579 & 0.0536 / 0.6022 \\
 & Question-Emb K-Means & - / - & 0.1098 / 0.2856 & 0.0859 / 0.3656 & 0.0615 / 0.5338 & 0.0522 / \textbf{0.6263} \\
 & k-Center Greedy & - / - & 0.1232 / 0.2197 & 0.0821 / 0.4036 & 0.0608 / 0.5355 & 0.0565 / 0.6151 \\
\rowcolor{orange!15} \cellcolor{white}  & \textbf{CoT-Feature (Ours)} & - / - & \textbf{0.1044} / \textbf{0.4334} & \textbf{0.0731} / \textbf{0.5839} & \textbf{0.0601} / \textbf{0.5646} & \textbf{0.0516} / 0.6013 \\
\cmidrule(lr){2-7}
& \multicolumn{6}{c}{\cellcolor{gray!15}\textit{Trainable Methods (Require Historical Data, N=25)}} \\
\rowcolor{blue!10} \cellcolor{white}  & Correctness K-Means & - / - & 0.0936 / 0.3125 & 0.0743 / 0.3435 & 0.0660 / 0.4498 & 0.0570 / 0.4770 \\
\rowcolor{blue!10} \cellcolor{white}  & IRT & - / - & 0.1054 / 0.2576 & 0.0723 / 0.3806 & 0.0552 / 0.6081 & 0.0513 / 0.6176 \\
\bottomrule
\end{tabular}
}
\end{table*}

\begin{table*}[t]
\centering
\caption{Performance comparison at $N=25$ using the GP-IRT estimator. Best performance among training-free methods is highlighted in \textbf{bold}. History-dependent methods are shown for reference only. Note that results at the 1\% budget for GPQA are omitted, as this strictly equates to a single instance, rendering coreset evaluation statistically meaningless.}
\label{table:25gpirt}
\resizebox{\textwidth}{!}{
\begin{tabular}{lcccccc}
\toprule
\multirow{2}{*}{\textbf{Benchmark}} & \multirow{2}{*}{\textbf{Method}} & \multicolumn{5}{c}{\textbf{Sampling Ratio}} \\
\cmidrule(lr){3-7}
& & \makecell{1\% \\ \scriptsize MAE ($\downarrow$) / Sim ($\uparrow$)} & \makecell{5\% \\ \scriptsize MAE ($\downarrow$) / Sim ($\uparrow$)} & \makecell{10\% \\ \scriptsize MAE ($\downarrow$) / Sim ($\uparrow$)} & \makecell{15\% \\ \scriptsize MAE ($\downarrow$) / Sim ($\uparrow$)} & \makecell{20\% \\ \scriptsize MAE ($\downarrow$) / Sim ($\uparrow$)} \\
\midrule
& \multicolumn{6}{c}{\cellcolor{gray!15}\textit{Training-Free Methods (Zero-Shot)}} \\
GSM8K & Random & 0.0827 / \textbf{0.8846} & 0.0297 / 0.9701 & 0.0202 / 0.9843 & \textbf{0.0158} / 0.9893 & 0.0131 / 0.9919 \\
 & Question-Emb K-Means & 0.0975 / 0.8443 & 0.0296 / \textbf{0.9715} & 0.0206 / 0.9776 & \textbf{0.0158} / 0.9866 & \textbf{0.0123} / 0.9928 \\
 & k-Center Greedy & 0.1016 / 0.8597 & \textbf{0.0289} / 0.9707 & 0.0216 / 0.9830 & 0.0166 / \textbf{0.9895} & 0.0136 / \textbf{0.9929} \\
\rowcolor{orange!15} \cellcolor{white}  & \textbf{CoT-Feature (Ours)} & \textbf{0.0800} / 0.8700 & 0.0290 / 0.9642 & \textbf{0.0197} / \textbf{0.9844} & 0.0166 / 0.9882 & 0.0135 / 0.9914 \\
\cmidrule(lr){2-7}
& \multicolumn{6}{c}{\cellcolor{gray!15}\textit{Trainable Methods (Require Historical Data, N=25)}} \\
\rowcolor{blue!10} \cellcolor{white}  & Correctness K-Means & 0.0830 / 0.8608 & 0.0281 / 0.9691 & 0.0183 / 0.9830 & 0.0156 / 0.9886 & 0.0121 / 0.9940 \\
\rowcolor{blue!10} \cellcolor{white}  & IRT & 0.0789 / 0.8957 & 0.0308 / 0.9659 & 0.0177 / 0.9838 & 0.0158 / 0.9877 & 0.0134 / 0.9909 \\
\midrule
& \multicolumn{6}{c}{\cellcolor{gray!15}\textit{Training-Free Methods (Zero-Shot)}} \\
MMLU & Random & 0.0212 / 0.9639 & 0.0116 / 0.9897 & 0.0089 / 0.9938 & 0.0072 / 0.9957 & 0.0060 / 0.9968 \\
 & Question-Emb K-Means & 0.0225 / 0.9606 & 0.0120 / 0.9891 & 0.0105 / 0.9933 & 0.0130 / 0.9947 & 0.0165 / 0.9961 \\
 & k-Center Greedy & \textbf{0.0198} / \textbf{0.9695} & 0.0190 / 0.9893 & 0.0191 / 0.9929 & 0.0209 / 0.9951 & 0.0205 / 0.9957 \\
\rowcolor{orange!15} \cellcolor{white}  & \textbf{CoT-Feature (Ours)} & 0.0223 / 0.9577 & \textbf{0.0111} / \textbf{0.9901} & \textbf{0.0076} / \textbf{0.9948} & \textbf{0.0068} / \textbf{0.9962} & \textbf{0.0059} / \textbf{0.9969} \\
\cmidrule(lr){2-7}
& \multicolumn{6}{c}{\cellcolor{gray!15}\textit{Trainable Methods (Require Historical Data, N=25)}} \\
\rowcolor{blue!10} \cellcolor{white}  & Correctness K-Means & 0.0222 / 0.9648 & 0.0142 / 0.9856 & 0.0121 / 0.9864 & 0.0113 / 0.9850 & 0.0110 / 0.9836 \\
\rowcolor{blue!10} \cellcolor{white}  & IRT & 0.0209 / 0.9651 & 0.0131 / 0.9883 & 0.0095 / 0.9932 & 0.0073 / 0.9961 & 0.0066 / 0.9968 \\
\midrule
& \multicolumn{6}{c}{\cellcolor{gray!15}\textit{Training-Free Methods (Zero-Shot)}} \\
MMLU-Pro & Random & 0.0250 / 0.9598 & 0.0104 / \textbf{0.9904} & 0.0074 / 0.9944 & 0.0060 / 0.9955 & 0.0051 / 0.9966 \\
 & Question-Emb K-Means & 0.0244 / 0.9547 & 0.0103 / 0.9895 & 0.0081 / \textbf{0.9949} & 0.0083 / 0.9955 & 0.0074 / 0.9968 \\
 & k-Center Greedy & 0.0240 / 0.9670 & 0.0146 / 0.9896 & 0.0149 / 0.9935 & 0.0162 / 0.9943 & 0.0162 / 0.9955 \\
\rowcolor{orange!15} \cellcolor{white}  & \textbf{CoT-Feature (Ours)} & \textbf{0.0240} / \textbf{0.9724} & \textbf{0.0102} / 0.9901 & \textbf{0.0069} / 0.9943 & \textbf{0.0058} / \textbf{0.9963} & \textbf{0.0047} / \textbf{0.9968} \\
\cmidrule(lr){2-7}
& \multicolumn{6}{c}{\cellcolor{gray!15}\textit{Trainable Methods (Require Historical Data, N=25)}} \\
\rowcolor{blue!10} \cellcolor{white}  & Correctness K-Means & 0.0234 / 0.9747 & 0.0091 / 0.9925 & 0.0081 / 0.9948 & 0.0075 / 0.9957 & 0.0070 / 0.9958 \\
\rowcolor{blue!10} \cellcolor{white}  & IRT & 0.0257 / 0.9549 & 0.0109 / 0.9903 & 0.0095 / 0.9943 & 0.0085 / 0.9960 & 0.0075 / 0.9971 \\
\midrule
& \multicolumn{6}{c}{\cellcolor{gray!15}\textit{Training-Free Methods (Zero-Shot)}} \\
GPQA & Random & - / - & 0.1019 / 0.2807 & 0.0679 / 0.4782 & 0.0568 / \textbf{0.5842} & 0.0497 / 0.6248 \\
 & Question-Emb K-Means & - / - & \textbf{0.0887} / 0.2759 & 0.0703 / 0.3900 & 0.0552 / 0.5704 & \textbf{0.0482} / \textbf{0.6388} \\
 & k-Center Greedy & - / - & 0.0973 / 0.2351 & 0.0701 / 0.4334 & 0.0551 / 0.5547 & 0.0519 / 0.6335 \\
\rowcolor{orange!15} \cellcolor{white}  & \textbf{CoT-Feature (Ours)} & - / - & 0.0922 / \textbf{0.5123} & \textbf{0.0649} / \textbf{0.6206} & \textbf{0.0545} / 0.5812 & 0.0482 / 0.6318 \\
\cmidrule(lr){2-7}
& \multicolumn{6}{c}{\cellcolor{gray!15}\textit{Trainable Methods (Require Historical Data, N=25)}} \\
\rowcolor{blue!10} \cellcolor{white}  & Correctness K-Means & - / - & 0.0728 / 0.3970 & 0.0597 / 0.4130 & 0.0559 / 0.5022 & 0.0503 / 0.5293 \\
\rowcolor{blue!10} \cellcolor{white}  & IRT & - / - & 0.0887 / 0.2545 & 0.0622 / 0.4248 & 0.0500 / 0.6402 & 0.0473 / 0.6475 \\
\bottomrule
\end{tabular}
}
\end{table*}

\textbf{Delineating Boundary Conditions via Task Complexity.} 
A cross-sectional analysis of Tables \ref{table:25sa} and \ref{table:25gpirt} delineates our approach's boundary conditions: \textit{CoT-Core's efficacy is intrinsically gated by benchmark difficulty.} On straightforward tasks like GSM8K, the method hits a \textbf{lower boundary (simplicity floor)}, yielding noticeable but bounded improvements. However, on expert-level benchmarks (MMLU-Pro, GPQA) where raw text acts as an information bottleneck, entering this \textbf{operative domain} via CoT-Core provides massive informational lift. Remarkably, our training-free approach bridges the gap with history-dependent methods. Under the advanced \textbf{GP-IRT} estimator, CoT-Core frequently matches or surpasses IRT and Correctness K-Means at larger budgets (e.g., $\ge 10\%$ on MMLU/MMLU-Pro), proving that capturing latent logical structure can serve as a highly effective alternative to compiling expensive historical evaluation logs.

\textbf{Robustness Across Estimators.}
Under the strictly training-free SA estimator (Table \ref{table:25sa}), CoT-Core's CoT-aware coresets demonstrate robust representativeness compared to Question-Emb K-Means and Random sampling across the majority of selection budgets. When transitioning to the advanced GP-IRT estimator (Table \ref{table:25gpirt}), the absolute predictive errors (MAE) naturally compress across all methods due to the estimator's parameterized advantage. Despite this tighter global margin, CoT-Core maintains highly competitive—and often leading—performance among training-free methods. Most notably, our approach showcases exceptional resilience in extreme high-compression regimes (e.g., $5\%$ to $10\%$), where it effectively preserves the relative performance ranking of evaluated models regardless of the scoring function. For instance, with a mere $5\%$ budget under GP-IRT, CoT-Core achieves a remarkable Ranking Similarity (Sim) of $0.5123$ on GPQA, nearly doubling the correlation preserved by traditional text clustering ($0.2759$), underscoring the structural stability of our reasoning-aware selection.

\subsection{Analysis}
\label{subsec:analysis}

\paragraph{The Simplicity Floor and Informational Boundaries.}
CoT-Core's performance gains are bounded by task complexity, directly reflecting the variance in \textit{discriminative information gain} introduced by the CoT trajectories. In straightforward benchmarks like GSM8K (our \textbf{lower boundary}), reasoning paths predominantly involve homogeneous basic arithmetic. Consequently, unrolled trajectories offer limited structural variance, and CoT verbosity occasionally introduces linguistic noise rather than useful signals. Conversely, reasoning-aware pruning finds its \textbf{robust operative domain} in complex benchmarks demanding diverse \textbf{underlying logical structures}. Invoking advanced theorems for STEM questions, or explicitly evaluating distractors for non-STEM questions, provides powerful spatial discriminators. Crucially, this informational advantage is highly resilient: even when the generator ($\mathcal{M}_{gen}$) hallucinates on excessively difficult questions, the specific \textit{nature} of its attempted retrieval or logical confusion provides a far more informative clustering signal than compressed raw question text.

\paragraph{Exposing Cognitive Isomorphism via Logical Trajectories}

\begin{figure*}[htbp]
    \centering
    \includegraphics[width=0.9\textwidth]{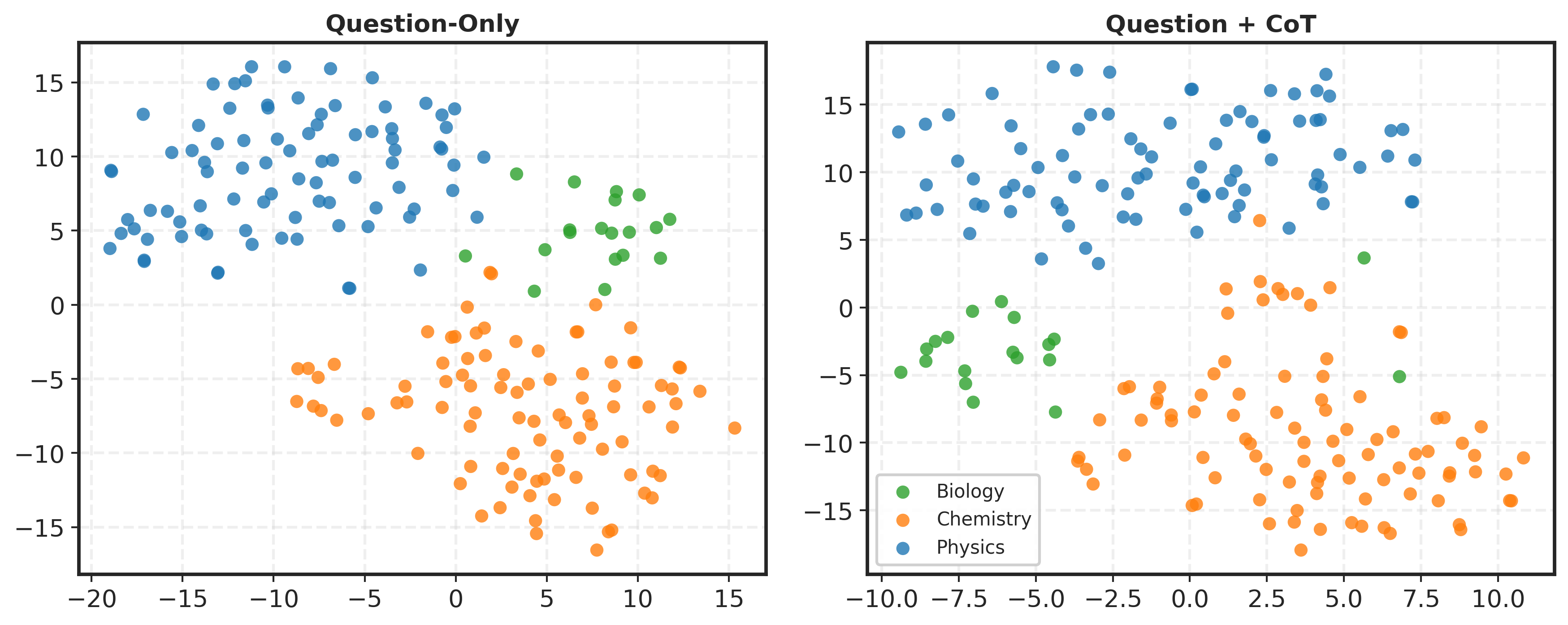}
    \caption{
        \textbf{t-SNE of GPQA embeddings.} (\textbf{Left}) Question-only embeddings are hindered by surface lexical traps and domain-specific vocabulary. (\textbf{Right}) CoT-Core reorganizes the latent space via logical isomorphisms, capturing underlying reasoning structures that transcend superficial text similarities.
    }
    \label{fig:gpqa_tsne}
\end{figure*}

To understand how CoT-Core externalizes implicit reasoning paths into explicit structural features within the representation space, we qualitatively analyze the reasoning manifold of the embeddings. We focus on GPQA, a highly complex benchmark spanning Physics, Chemistry, and Biology, as it provides an ideal testbed for observing cross-disciplinary semantic shifts.

\textbf{Macro-Level Manifold: Breaking Lexical Traps.} 
Figure \ref{fig:gpqa_tsne} visualizes the embedding spaces before and after introducing Chain-of-Thought (CoT) trajectories. In the Question-Only space (left), the dense retriever naturally separates the three sciences into distinct clusters. However, this apparent separability is largely driven by \textit{lexical traps}. Because domain-specific vocabularies (e.g., "cell" in Biology vs. "velocity" in Physics) are highly disjoint, the encoder creates artificial boundaries based on surface-level entities rather than intrinsic logical structures. 

Conversely, the Question+CoT space (right) exhibits a fascinating phenomenon: while maintaining global structure, we observe deliberate inter-cluster merging. Specifically, the originally monolithic domain clusters (e.g., Biology, in green) begin to fragment, with individual instances clearly breaking away from their main group. Rather than being noise, this structural dispersion indicates that CoT-Core forces the embedding space to align with the underlying reasoning structures. Questions within the same domain that require fundamentally different logical trajectories (e.g., a biology question requiring mathematical calculation versus one requiring factual recall) are pulled apart, disrupting the artificial, lexicon-driven boundaries.

\textbf{Micro-Level Case Study: Correcting the Logical Manifold.} 
To validate that spatial reorganization captures underlying reasoning rather than surface semantics, we examine two contrasting GPQA scenarios (detailed in Appendix \ref{sec:appendix_case_study}).

\textit{Case 1: Uncovering Logical Isomorphism (False-Negative Correction).} 
Two physics problems---one regarding a CERN Bubble Chamber and another on a spherical detector---exhibit a high Question-Only cosine distance of 0.5461 due to disjoint vocabularies. However, CoT unrolling reveals an identical mathematical skeleton (relativistic kinematics and time dilation). By encoding this shared logic, CoT-Core collapses their distance by 61\% to 0.2128, ensuring clustering targets reasoning skills rather than experimental dressing.

\textit{Case 2: Escaping Lexical Traps (False-Positive Correction).} 
Conversely, traditional encoders frequently conflate distinct problems sharing deceptive surface vocabulary. Two astrophysics questions about a $6000\text{K}$ spotted star (quantum state population vs. macroscopic photometric transit) are dangerously entangled with a distance of 0.1957. CoT-Core exposes their divergent physical laws (microscopic Boltzmann distribution vs. macroscopic Stefan-Boltzmann law), shattering the lexical illusion and nearly doubling their distance to 0.3890 to prevent pruning non-redundant knowledge.

\subsection{Ablation Studies}
\label{subsec:ablation}

\noindent\textbf{Impact of the CoT Generator: The Verbosity Penalty.}
We investigate the impact of the LLM used to externalize reasoning trajectories on MMLU-Pro, varying the generator from lightweight (1B) to frontier models (72B) while fixing the encoder to BGE-M3. 

\begin{wraptable}{r}{0.48\textwidth}
    \vspace{-1.5em}
    \centering
    \caption{Ablation on generator scale (MMLU-Pro). Lightweight models perform remarkably competitively, particularly at extreme compression ratios.}
    \label{tab:ablation_generator}
    \resizebox{\linewidth}{!}{%
    \begin{tabular}{lccc}
    \toprule
    \textbf{Generator} & \textbf{1\% Budget} & \textbf{10\% Budget} & \textbf{20\% Budget} \\
    \midrule
    LLaMA-3-1B & $0.0280 \pm 0.0204$ & $0.0109 \pm 0.0082$ & $0.0057 \pm 0.0043$ \\
    Mistral-8B & $0.0316 \pm 0.0240$ & $0.0081 \pm 0.0058$ & $0.0066 \pm 0.0051$ \\
    Phi-4 & $0.0295 \pm 0.0241$ & $0.0082 \pm 0.0061$ & $\mathbf{0.0054 \pm 0.0043}$ \\
    Qwen-72B & $0.0305 \pm 0.0236$ & $\mathbf{0.0077 \pm 0.0060}$ & $0.0065 \pm 0.0048$ \\
    \bottomrule
    \end{tabular}%
    }
    \vspace{-1em}
\end{wraptable}

A counter-intuitive observation is that scaling to a massive 72B generator yields marginal, often statistically insignificant, clustering improvements over lightweight alternatives. This highlights a core premise of our framework: \textbf{trajectory correctness is secondary to logical structure}. Even when 1B models derive incorrect final answers, they successfully externalize the requisite reasoning steps, fully exposing the problem's underlying reasoning manifold. Furthermore, frontier models suffer from a \textbf{Verbosity Penalty}. Extensive instruction tuning causes 72B models to generate excessive conversational boilerplate. Because dense encoders are sensitive to lexical frequency \citep{shi2023large}, this universal linguistic noise artificially inflates similarities between fundamentally disparate problems. Consequently, the compact, raw logical trajectories of smaller models prove highly robust for geometric clustering, confirming CoT-Core's cost-efficiency.

\noindent\textbf{Impact of Hierarchical Cleaning: The Optimal Abstraction Horizon.}
To isolate CoT-Core's mechanisms, we evaluate three abstraction levels on GPQA using the Phi-4 generator (prompts in Appendix \ref{sec:appendix_prompts}): \textit{Raw CoT} (unaltered), \textit{Cleaned CoT} (conversational boilerplate removed, entities retained), and \textit{Abstract CoT} (extreme compression retaining only pure logical structures).

\begin{wraptable}{r}{0.48\textwidth}
    \vspace{-1.5em}
    \centering
    \caption{Ablation on Hierarchical Cleaning (GPQA). \textit{Cleaned CoT} achieves the optimal balance.}
    \label{tab:ablation_clean}
    \resizebox{\linewidth}{!}{%
    \begin{tabular}{lccc}
    \toprule
    \multirow{2}{*}{\textbf{Abstraction Level}} & \multicolumn{3}{c}{\textbf{GPQA Budget (MAE $\downarrow$ / Sim $\uparrow$)}} \\
    \cmidrule(lr){2-4}
    & \textbf{5\%} & \textbf{10\%} & \textbf{20\%} \\
    \midrule
    Raw CoT & 0.1044 / 0.4334 & 0.0731 / \textbf{0.5839} & 0.0516 / 0.6013 \\
    Cleaned CoT & \textbf{0.1007} / \textbf{0.5041} & \textbf{0.0666} / 0.4970 & \textbf{0.0442} / \textbf{0.6438} \\
    Abstract CoT & 0.1206 / 0.3561 & 0.0831 / 0.4506 & 0.0518 / 0.5714 \\
    \bottomrule
    \end{tabular}%
    }
    \vspace{-1em}
\end{wraptable}

As Table \ref{tab:ablation_clean} shows, the intermediate \textit{Cleaned CoT} consistently minimizes predictive error. Stripping universally shared boilerplate drastically reduces linguistic noise, sharpening manifold boundaries. However, extreme compression (\textit{Abstract CoT}) severely degrades performance, exposing a critical failure mode: \textbf{Semantic Starvation}. Unlike elementary math, expert domains require domain-specific semantic anchors to contextualize the underlying logic. Stripping these disciplinary groundings deprives the encoder of essential contextual dimensionality. Thus, CoT-Core operates optimally by pruning generic noise while preserving the semantic anchors required to contextualize the latent logical structure.

\section{Conclusion}
\label{sec:conclusion}

We presented CoT-Core, a training-free coreset framework that circumvents surface lexical traps by clustering the underlying reasoning manifold of unrolled Chain-of-Thought trajectories, successfully aligning lexically disparate problems via their intrinsic logical isomorphism. Extensive evaluations confirm CoT-Core consistently yields high-fidelity proxy estimations for both standard and advanced probabilistic estimators. Crucially, we delineate the boundary conditions of reasoning-aware pruning: its efficacy intrinsically scales with task complexity, bounded by a simplicity floor on basic arithmetic while delivering massive informational gains on expert-level benchmarks. Furthermore, achieving optimal representation requires balancing model-induced conversational noise (\textit{Verbosity Penalty}) against the severe loss of disciplinary context (\textit{Semantic Starvation}). Ultimately, CoT-Core provides a scalable, data-free alternative to compiling expensive historical evaluation logs, proving that capturing latent logical structures is sufficient for robust benchmark compression and accelerating continuous LLM development.

\bibliography{colm2026_conference}
\bibliographystyle{colm2026_conference}

\clearpage 
\appendix


\section{Implementation Details}
\label{sec:appendix_implementation}
This section provides comprehensive details regarding our experimental setup to facilitate reproducibility.

\subsection{Model Configurations}
We employ the pre-trained dense encoder (e.g., BGE-M3) for contextualized trajectory embedding. For generating the Unrolled Reasoning Trajectories, we utilize the generator $\mathcal{M}_{gen}$ with the following generation hyperparameters: temperature $T=0.7$, and maximum token length set to 1024.

\subsection{Prompt Templates for Hierarchical Cleaning}
\label{sec:appendix_prompts}

To elicit implicit reasoning from the generator and subsequently perform our hierarchical cleaning ablations (as discussed in Section \ref{subsec:ablation}), we utilize specific system prompts. To strictly control for confounding variables, both the initial reasoning generation and the subsequent hierarchical cleaning steps employ the identical Phi-4 model. 

For the initial generation, we append a standard CoT trigger $p_{cot}$ (e.g., \textit{``Let's think step by step''}) to the original question $q_i$. The subsequent abstraction processes utilize the exact prompt templates presented below.

\vspace{1em}

\begin{tcolorbox}[
    colback=gray!5, 
    colframe=gray!60!black, 
    title=\textbf{System Prompt: Level-1 Denoised (Cleaned CoT)}, 
    fonttitle=\bfseries\sffamily, 
    arc=3mm, 
    boxrule=0.8pt, 
    left=6pt, right=6pt, top=6pt, bottom=6pt
]
You are a strict reasoning text cleaner. Your task is to extract only the valid, successful logical steps from the provided reasoning process that lead to the final answer. Strictly remove any self-corrections, dead-ends, hesitations, or redundant thoughts. Retain all necessary calculations, entities, and the core logical trajectory. Output ONLY the cleaned reasoning text without any introductory, conversational, or concluding remarks.
\end{tcolorbox}

\vspace{1em}

\begin{tcolorbox}[
    colback=gray!5, 
    colframe=gray!60!black, 
    title=\textbf{System Prompt: Level-2 Skeleton (Abstract CoT)}, 
    fonttitle=\bfseries\sffamily, 
    arc=3mm, 
    boxrule=0.8pt, 
    left=6pt, right=6pt, top=6pt, bottom=6pt
]
You are an expert in mathematical and logical abstraction. Your task is to extract the 'Logical Skeleton' (Granularity 2) from the provided Question and Reasoning. \\
Strictly follow these rules: \\
1. Abstract away ALL specific numbers, values, and concrete entities (e.g., replace '3 apples' with 'initial quantity', '18 km/h' with 'velocity', 'John' with 'agent'). \\
2. Summarize the problem setup and the step-by-step reasoning process as a sequence of generalized meta-operations (e.g., 'Identify variables $\rightarrow$ Set up system of linear equations $\rightarrow$ Solve by substitution'). \\
3. Preserve the exact logical trajectory, sequence of steps, and causal relationships. \\
4. Output ONLY the abstracted step-by-step sequence. Do not include any conversational filler, introductory, or concluding remarks.
\end{tcolorbox}

\section{Case Study: Full Problem Descriptions and Reasoning Trajectories}
\label{sec:appendix_case_study}

This appendix provides the complete original text for the four GPQA questions analyzed in our micro-level case study (Section \ref{subsec:results}), alongside their unrolled Chain-of-Thought (CoT) trajectories generated by the $\mathcal{M}_{gen}$ (Phi-4) model.

\subsection{Case 1: Uncovering Logical Isomorphism (False-Negative Correction)}
These two problems are lexically disjoint but share an identical underlying mathematical skeleton rooted in relativistic kinematics and time dilation.

\paragraph{Question A (Experimental Resolution):}
In the CERN Bubble Chamber a decay occurs, $X^{0}\rightarrow Y^{+}Z^{-}$ in $\tau_{0}=8\times10^{-16}$ s, i.e. the proper lifetime of $X^{0}$. What minimum resolution is needed to observe at least 30\% of the decays? Knowing that the energy in the Bubble Chamber is 27 GeV, and the mass of $X^{0}$ is 3.41 GeV.

\begin{cotbox}[Unrolled CoT Trajectory (Question A)]
To solve this problem, we need to determine the minimum resolution required to observe at least 30\% of the decays of the particle $X^0$. This involves understanding the relationship between the proper lifetime, the energy of the particle, and the distance it travels before decaying.

\textbf{Step 1: Calculate the Lorentz factor ($\gamma$)}

The Lorentz factor $\gamma$ is given by:
$$\gamma = \frac{E}{m c^2}$$
where:
\begin{itemize}
    \item $E = 27 \text{ GeV}$ is the total energy,
    \item $m = 3.41 \text{ GeV}/c^2$ is the rest mass of $X^0$,
    \item $c$ is the speed of light.
\end{itemize}

Substituting the values:
$$\gamma = \frac{27 \text{ GeV}}{3.41 \text{ GeV}} \approx 7.92$$

\textbf{Step 2: Calculate the dilated lifetime ($\tau$)}

The dilated lifetime $\tau$ is given by:
$$\tau = \gamma \tau_0$$
where $\tau_0 = 8 \times 10^{-16} \text{ s}$ is the proper lifetime.

$$\tau = 7.92 \times 8 \times 10^{-16} \text{ s} \approx 6.336 \times 10^{-15} \text{ s}$$

\textbf{Step 3: Calculate the average distance traveled before decay}

The average distance $d$ traveled by the particle before decaying is:
$$d = v \tau$$
where $v \approx c$ for high-energy particles. Thus:
$$d \approx c \tau = (3 \times 10^8 \text{ m/s}) \times 6.336 \times 10^{-15} \text{ s} \approx 1.9008 \times 10^{-6} \text{ m}$$

\textbf{Step 4: Determine the resolution needed to observe 30\% of decays}

To observe at least 30\% of the decays, the resolution $\Delta x$ should be such that:
$$\Delta x \leq \frac{d}{\ln(1/0.3)}$$

Calculating $\ln(1/0.3)$:
$$\ln(1/0.3) \approx 1.204$$

Thus:
$$\Delta x \leq \frac{1.9008 \times 10^{-6} \text{ m}}{1.204} \approx 1.578 \times 10^{-6} \text{ m}$$

\textbf{Step 5: Compare with given options}

The closest option that satisfies $\Delta x \leq 1.578 \times 10^{-6} \text{ m}$ is:
C) $2.08 \times 10^{-6} \text{ m}$

Thus, the minimum resolution needed is approximately $2.08 \times 10^{-6} \text{ m}$.

\textbf{ANSWER:} C
\end{cotbox}

\paragraph{Question B (Kinematic State):}
Particles are collided at the center of a spherical detector producing new type of particles that travel uninterrupted at ultra-relativistic velocities highly centered around Lorentz factor of $\sim 20$. On average, one third of these fast-decaying particles reaches the detector inner walls. The radius of the detector is 30 meters. What Lorentz factor is needed in order to have about two thirds of these particles reaching the detector inner walls?

\begin{cotbox}[Unrolled CoT Trajectory (Question B)]
To solve this problem, we need to understand the relationship between the Lorentz factor ($\gamma$) and the distance traveled by the particles before they decay. The Lorentz factor is given by:

$$\gamma = \frac{1}{\sqrt{1 - v^2/c^2}}$$

where $v$ is the velocity of the particle and $c$ is the speed of light.

The mean lifetime of a particle in its rest frame is $\tau_0$. In the lab frame, due to time dilation, the mean lifetime is $\tau = \gamma \tau_0$.

The average distance $d$ a particle travels before decaying is given by:

$$d = v \tau = v \gamma \tau_0$$

For ultra-relativistic particles, $v \approx c$, so:

$$d \approx c \gamma \tau_0$$

We know that with $\gamma \approx 20$, one third of the particles reach the detector's inner walls, which are 30 meters away. This implies that the average distance traveled by these particles is approximately 30 meters.

Now, we want two thirds of the particles to reach the detector, meaning the average distance should be approximately 60 meters (since two thirds reaching implies the average distance is twice as much as when one third reaches).

Using the proportionality:

$$\frac{d_2}{d_1} = \frac{\gamma_2}{\gamma_1}$$

where $d_1 = 30$ meters and $d_2 = 60$ meters, and $\gamma_1 = 20$, we solve for $\gamma_2$:

$$\frac{60}{30} = \frac{\gamma_2}{20}$$

$$2 = \frac{\gamma_2}{20}$$

$$\gamma_2 = 40$$

Thus, the Lorentz factor needed for about two thirds of the particles to reach the detector's inner walls is 40.

ANSWER: C
\end{cotbox}

\paragraph{Analysis of the Isomorphism:}
This case illustrates the False-Negative problem inherent in standard text-embedding approaches and how unrolled Chain-of-Thought (CoT) trajectories mitigate it. 

If we rely solely on the problem formulations, Question A and Question B appear semantically disjoint. Question A is framed as an experimental high-energy physics problem (featuring entities like ``CERN Bubble Chamber,'' ``27 GeV,'' and ``resolution''), whereas Question B is framed as a macroscopic kinematic problem (featuring ``spherical detector,'' ``30 meters,'' and ``inner walls''). A naive embedding model would map these two questions to distant regions in the latent space due to their lexical divergence.

However, the unrolled CoTs reveal a shared, identical mathematical skeleton. Both reasoning trajectories sequentially invoke the same underlying physical principles:
\begin{enumerate}
    \item \textbf{Relativistic Kinematics:} Calculating or utilizing the Lorentz factor ($\gamma$).
    \item \textbf{Time Dilation:} Linking proper lifetime to dilated lifetime ($\tau = \gamma \tau_0$).
    \item \textbf{Mean Decay Length:} Formulating the distance traveled before decay ($d \approx c \gamma \tau_0$).
    \item \textbf{Threshold Probability:} Correlating the traveled distance with the survival/decay ratio.
\end{enumerate}

By exposing this implicit reasoning structure, the CoT acts as a semantic equalizer. The generation of shared specialized tokens (e.g., $\gamma$, $\tau_0$, $v \approx c$) and structurally identical mathematical equations forces the embedding space to recognize their deep logical isomorphism. Consequently, these two instances, which would be falsely separated (False-Negative) in a purely lexical feature space, are successfully clustered together based on their true cognitive and reasoning demands.\footnote{It is worth noting that the LLM exhibits a minor reasoning hallucination in Question B (applying a linear proportion to exponential decay). However, this precisely highlights the robustness of our approach: the embedding alignment relies on the \textit{structural and conceptual vocabulary} (Lorentz transformations, time dilation) extracted by the CoT, rather than the strict arithmetic correctness of the final step.}

\subsection{Case 2: Escaping Lexical Traps (False-Positive Correction)}
These two problems share extensive surface-level vocabulary regarding spotted stars and effective temperatures, but require fundamentally orthogonal physical laws (quantum statistical mechanics vs. macroscopic photometric geometry).

\paragraph{Question C (Quantum States):}
Astronomers are studying a star with a 1.5 solar radius and 1.1 solar masses. When the star's surface is not covered by dark spots, its $T_{\text{eff}}$ is 6000K. However, when 40\% of its surface is covered by spots, the overall photospheric effective temperature decreases to 5500K. In the stellar photosphere, when examining the ratio of the number of neutral atoms of Ti in two energetic levels (level 1 and level 2), astronomers have observed that this ratio decreases when the star has spots. What is the factor by which this ratio changes when the star does not have spots compared to when it has spots? Note that the transition between the energy levels under consideration corresponds to a wavelength of approximately 1448 \AA. Assume that the stellar photosphere is in LTE.

\begin{cotbox}[Unrolled CoT Trajectory (Question C)]
To solve this problem, we need to understand how the ratio of the number of neutral titanium (Ti) atoms in two energy levels changes with temperature. This is governed by the Boltzmann distribution, which describes the population of atoms in different energy states at thermal equilibrium.

The Boltzmann distribution is given by:

$$\frac{n_2}{n_1} = \frac{g_2}{g_1} \exp\left(-\frac{\Delta E}{kT}\right)$$

where:
- $n_2$ and $n_1$ are the number of atoms in energy levels 2 and 1, respectively.
- $g_2$ and $g_1$ are the statistical weights (degeneracies) of the energy levels.
- $\Delta E$ is the energy difference between the two levels.
- $k$ is the Boltzmann constant.
- $T$ is the temperature.

The energy difference $\Delta E$ can be related to the wavelength $\lambda$ of the transition using the formula:

$$\Delta E = \frac{hc}{\lambda}$$

where:
- $h$ is Planck's constant.
- $c$ is the speed of light.

Given:
- $\lambda = 1448 \, \text{Å} = 1448 \times 10^{-10} \, \text{m}$
- $T_1 = 6000 \, \text{K}$ (without spots)
- $T_2 = 5500 \, \text{K}$ (with spots)

We need to find the factor by which the ratio $\frac{n_2}{n_1}$ changes from $T_1$ to $T_2$.

The ratio of the populations at two different temperatures is:

$$\frac{\left(\frac{n_2}{n_1}\right)_{T_1}}{\left(\frac{n_2}{n_1}\right)_{T_2}} = \frac{\exp\left(-\frac{\Delta E}{kT_1}\right)}{\exp\left(-\frac{\Delta E}{kT_2}\right)} = \exp\left(\frac{\Delta E}{k}\left(\frac{1}{T_2} - \frac{1}{T_1}\right)\right)$$

Substitute $\Delta E = \frac{hc}{\lambda}$:

$$\frac{\left(\frac{n_2}{n_1}\right)_{T_1}}{\left(\frac{n_2}{n_1}\right)_{T_2}} = \exp\left(\frac{hc}{\lambda k}\left(\frac{1}{T_2} - \frac{1}{T_1}\right)\right)$$

Now, calculate the numerical value:

- $h = 6.626 \times 10^{-34} \, \text{J s}$
- $c = 3 \times 10^8 \, \text{m/s}$
- $k = 1.381 \times 10^{-23} \, \text{J/K}$

$$\frac{hc}{\lambda k} = \frac{(6.626 \times 10^{-34})(3 \times 10^8)}{1448 \times 10^{-10} \times 1.381 \times 10^{-23}}$$

$$= \frac{1.9878 \times 10^{-25}}{1.999688 \times 10^{-26}} \approx 9.94 \times 10^0$$

Now, calculate the temperature difference term:

$$\frac{1}{T_2} - \frac{1}{T_1} = \frac{1}{5500} - \frac{1}{6000} = \frac{6000 - 5500}{5500 \times 6000} = \frac{500}{33000000} \approx 1.515 \times 10^{-5}$$

Now, calculate the exponential:

$$\exp\left(9.94 \times 1.515 \times 10^{-5}\right) = \exp(0.0001506) \approx 1.1507$$

Thus, the factor by which the ratio changes is approximately 1.15, which is closest to option A.

ANSWER: A
\end{cotbox}

\paragraph{Question D (Photometric Flux):}
Astronomers are currently observing a star with a radius equal to that of the Sun. One hemisphere of the star is covered in dark spots with a filling factor of 20\%. The star has an effective temperature ($T_{\text{eff}}$) of 6000K, and the spots exhibit a temperature difference of 1000K. As only one hemisphere is spotty, photometric time-series observations will reveal periodic variations in brightness due to rotational modulation. Interestingly, this situation can closely resemble the presence of an exoplanet. To produce the same amplitude signal in the star's light curve (if the star was not covered by spots!), what should be the radius of a hypothetical exoplanet relative to the radius of the host star (i.e. $R_{\text{pl}}/R_{\text{star}}$)?

\begin{cotbox}[Unrolled CoT Trajectory (Question D)]
To solve this problem, we need to determine the radius of a hypothetical exoplanet that would produce the same amplitude of brightness variation in the star's light curve as the dark spots on one hemisphere of the star.

\textbf{Step-by-step Solution:}

1. \textbf{Understand the Spot Coverage:}
   - The star has a filling factor of 20\% for the dark spots on one hemisphere. This means 20\% of the visible hemisphere is covered by spots.

2. \textbf{Calculate the Area of the Spots:}
   - The visible hemisphere of the star has an area of $2\pi R_{\text{star}}^2$.
   - The area covered by spots is $0.2 \times 2\pi R_{\text{star}}^2 = 0.4\pi R_{\text{star}}^2$.

3. \textbf{Brightness Reduction Due to Spots:}
   - The effective temperature of the star is $T_{\text{eff}} = 6000 \, \text{K}$.
   - The temperature of the spots is $T_{\text{spot}} = 6000 \, \text{K} - 1000 \, \text{K} = 5000 \, \text{K}$.

4. \textbf{Calculate the Flux Reduction:}
   - The flux from a blackbody is proportional to $T^4$.
   - The flux reduction factor for the spots is $\left(\frac{5000}{6000}\right)^4 = \left(\frac{5}{6}\right)^4$.

5. \textbf{Calculate the Total Flux Reduction:}
   - The total flux reduction due to the spots is the product of the area covered by spots and the flux reduction factor:
   $$\Delta F_{\text{spots}} = 0.4\pi R_{\text{star}}^2 \times \left(\frac{5}{6}\right)^4$$

6. \textbf{Calculate the Equivalent Exoplanet Radius:}
   - The flux reduction due to a transiting exoplanet is given by the ratio of the area of the planet to the area of the star:
   $$\Delta F_{\text{planet}} = \pi R_{\text{planet}}^2 / \pi R_{\text{star}}^2 = (R_{\text{planet}}/R_{\text{star}})^2$$
   - Set $\Delta F_{\text{spots}} = \Delta F_{\text{planet}}$:
   $$0.4 \times \left(\frac{5}{6}\right)^4 = (R_{\text{planet}}/R_{\text{star}})^2$$

7. \textbf{Solve for $R_{\text{planet}}/R_{\text{star}}$:}
   - Calculate $\left(\frac{5}{6}\right)^4$:
   $$\left(\frac{5}{6}\right)^4 = \frac{625}{1296} \approx 0.4823$$
   - Substitute back:
   $$0.4 \times 0.4823 \approx 0.1929$$
   - Solve for $R_{\text{planet}}/R_{\text{star}}$:
   $$(R_{\text{planet}}/R_{\text{star}})^2 = 0.1929$$
   $$R_{\text{planet}}/R_{\text{star}} = \sqrt{0.1929} \approx 0.439$$

8. \textbf{Choose the Closest Answer:}
   - The closest answer to 0.439 is 0.39.

ANSWER: D
\end{cotbox}

\paragraph{Analysis of the Orthogonality:}
This case demonstrates the ``Lexical Trap'' (False-Positive) problem in benchmark evaluation and how unrolled CoT prevents erroneous clustering. 

At the surface level, Question C and Question D are lexically entangled. Both problem descriptions share a high density of domain-specific keywords and identical numerical setups: ``star,'' ``dark spots,'' ``effective temperature,'' and ``6000K.'' A naive text-embedding model, which heavily weights term frequencies and surface-level semantic proximity, would inevitably map these two questions into the same cluster, classifying them as highly similar stellar astrophysics problems.

However, the unrolled CoT trajectories expose fundamentally orthogonal underlying mathematical and physical engines:
\begin{itemize}
    \item \textbf{Question C (Quantum Statistical Mechanics):} The reasoning trajectory is entirely driven by atomic physics. It invokes the \textit{Boltzmann distribution} to calculate microscopic population states, generating specialized tokens like energy differences ($\Delta E$), Planck's constant ($h$), and exponential decay functions dependent on temperature ($\exp(-\Delta E/kT)$).
    \item \textbf{Question D (Macroscopic Photometry):} The reasoning trajectory operates on geometric and thermodynamic scales. It relies on the \textit{Stefan-Boltzmann law} (where flux is proportional to $T^4$) and geometrical area ratios (projected disk areas and filling factors) to calculate macroscopic light curve variations.
\end{itemize}

By expanding the implicit reasoning paths, the CoT introduces dimensionally distinct vocabularies (e.g., $hc/\lambda k$ vs. $T^4$ ratios). This forced articulation of the underlying physical laws acts as a semantic repeller in the latent space, effectively pushing these two lexically similar but logically distinct problems apart, thereby resolving the False-Positive collision.\footnote{It should be noted that in Question D, the LLM utilizes a simplified geometric assumption regarding the spot area on a sphere versus a projected cross-sectional disk (ignoring limb darkening effects). Nevertheless, this pedagogical simplification does not hinder the embedding process. The core contribution of the CoT remains intact: successfully shifting the semantic representation from the misleading lexical surface to the correct physical domain (macroscopic geometric flux rather than quantum states).}

\section{Extended Main Results}
\label{sec:appendix_extended_results}
In the main text (Section \ref{subsec:results}), we reported the performance under the low-resource setting ($N=25$). Here, we present the full evaluation results across larger historical budgets $N \in \{50, 100, 200\}$.

\begin{table}[H]
\centering
\caption{Performance comparison at N=50 using SA proxy. Best performance among Training-Free methods is highlighted in \textbf{bold}. Trainable methods are shown for reference only.}
\resizebox{\textwidth}{!}{
\begin{tabular}{lcccccc}
\toprule
\multirow{2}{*}{\textbf{Benchmark}} & \multirow{2}{*}{\textbf{Method}} & \multicolumn{5}{c}{\textbf{Sampling Ratio}} \\
\cmidrule(lr){3-7}
& & \makecell{1\% \\ \scriptsize MAE ($\downarrow$) / Sim ($\uparrow$)} & \makecell{5\% \\ \scriptsize MAE ($\downarrow$) / Sim ($\uparrow$)} & \makecell{10\% \\ \scriptsize MAE ($\downarrow$) / Sim ($\uparrow$)} & \makecell{15\% \\ \scriptsize MAE ($\downarrow$) / Sim ($\uparrow$)} & \makecell{20\% \\ \scriptsize MAE ($\downarrow$) / Sim ($\uparrow$)} \\
\midrule
& \multicolumn{6}{c}{\cellcolor{gray!15}\textit{Training-Free Methods (Zero-Shot)}} \\
GSM8K & Random & 0.0890 / 0.9044 & 0.0382 / \textbf{0.9747} & 0.0246 / 0.9851 & 0.0187 / \textbf{0.9897} & 0.0142 / 0.9920 \\
 & Question-Emb K-Means & 0.0875 / 0.8951 & \textbf{0.0321} / 0.9743 & 0.0241 / 0.9797 & \textbf{0.0175} / 0.9871 & \textbf{0.0133} / \textbf{0.9931} \\
 & k-Center Greedy & \textbf{0.0683} / \textbf{0.9118} & 0.0349 / 0.9738 & 0.0283 / 0.9847 & 0.0218 / \textbf{0.9897} & 0.0175 / 0.9925 \\
\rowcolor{orange!15} \cellcolor{white}  & \textbf{CoT-Feature (Ours)} & 0.0755 / 0.8884 & 0.0331 / 0.9696 & \textbf{0.0209} / \textbf{0.9853} & 0.0197 / 0.9884 & 0.0154 / 0.9918 \\
\cmidrule(lr){2-7}
& \multicolumn{6}{c}{\cellcolor{gray!15}\textit{Trainable Methods (Require Historical Data, N=50)}} \\
\rowcolor{blue!10} \cellcolor{white}  & Correctness K-Means & 0.0623 / 0.9130 & 0.0339 / 0.9748 & 0.0256 / 0.9816 & 0.0230 / 0.9848 & 0.0195 / 0.9887 \\
\rowcolor{blue!10} \cellcolor{white}  & IRT & 0.0670 / 0.9053 & 0.0329 / 0.9754 & 0.0241 / 0.9833 & 0.0169 / 0.9908 & 0.0143 / 0.9907 \\
\midrule
& \multicolumn{6}{c}{\cellcolor{gray!15}\textit{Training-Free Methods (Zero-Shot)}} \\
MMLU & Random & 0.0330 / \textbf{0.9489} & 0.0149 / 0.9880 & 0.0102 / 0.9935 & \textbf{0.0079} / 0.9957 & 0.0063 / 0.9969 \\
 & Question-Emb K-Means & \textbf{0.0298} / 0.9484 & 0.0215 / 0.9877 & 0.0154 / 0.9923 & 0.0203 / 0.9943 & 0.0252 / 0.9958 \\
 & k-Center Greedy & 0.0477 / 0.9461 & 0.0530 / 0.9867 & 0.0393 / 0.9909 & 0.0360 / 0.9942 & 0.0317 / 0.9952 \\
\rowcolor{orange!15} \cellcolor{white}  & \textbf{CoT-Feature (Ours)} & 0.0324 / 0.9353 & \textbf{0.0118} / \textbf{0.9881} & \textbf{0.0081} / \textbf{0.9941} & 0.0081 / \textbf{0.9964} & \textbf{0.0060} / \textbf{0.9971} \\
\cmidrule(lr){2-7}
& \multicolumn{6}{c}{\cellcolor{gray!15}\textit{Trainable Methods (Require Historical Data, N=50)}} \\
\rowcolor{blue!10} \cellcolor{white}  & Correctness K-Means & 0.0272 / 0.9638 & 0.0191 / 0.9789 & 0.0161 / 0.9825 & 0.0152 / 0.9838 & 0.0144 / 0.9851 \\
\rowcolor{blue!10} \cellcolor{white}  & IRT & 0.0225 / 0.9588 & 0.0113 / 0.9896 & 0.0083 / 0.9938 & 0.0069 / 0.9960 & 0.0065 / 0.9968 \\
\midrule
& \multicolumn{6}{c}{\cellcolor{gray!15}\textit{Training-Free Methods (Zero-Shot)}} \\
MMLU-Pro & Random & 0.0311 / 0.9556 & 0.0132 / \textbf{0.9899} & 0.0093 / 0.9941 & \textbf{0.0070} / 0.9956 & 0.0058 / 0.9967 \\
 & Question-Emb K-Means & 0.0300 / 0.9579 & 0.0133 / 0.9890 & 0.0111 / \textbf{0.9947} & 0.0116 / 0.9957 & 0.0100 / \textbf{0.9969} \\
 & k-Center Greedy & 0.0369 / 0.9524 & 0.0432 / 0.9857 & 0.0334 / 0.9923 & 0.0306 / 0.9936 & 0.0272 / 0.9957 \\
\rowcolor{orange!15} \cellcolor{white}  & \textbf{CoT-Feature (Ours)} & \textbf{0.0296} / \textbf{0.9682} & \textbf{0.0124} / 0.9893 & \textbf{0.0082} / 0.9941 & 0.0072 / \textbf{0.9965} & \textbf{0.0055} / \textbf{0.9969} \\
\cmidrule(lr){2-7}
& \multicolumn{6}{c}{\cellcolor{gray!15}\textit{Trainable Methods (Require Historical Data, N=50)}} \\
\rowcolor{blue!10} \cellcolor{white}  & Correctness K-Means & 0.0283 / 0.9769 & 0.0172 / 0.9903 & 0.0152 / 0.9933 & 0.0131 / 0.9942 & 0.0113 / 0.9947 \\
\rowcolor{blue!10} \cellcolor{white}  & IRT & 0.0412 / 0.9642 & 0.0208 / 0.9902 & 0.0151 / 0.9938 & 0.0122 / 0.9955 & 0.0102 / 0.9963 \\
\midrule
& \multicolumn{6}{c}{\cellcolor{gray!15}\textit{Training-Free Methods (Zero-Shot)}} \\
GPQA & Random & - / - & 0.1252 / 0.2851 & 0.0797 / 0.4480 & 0.0635 / 0.5579 & 0.0536 / 0.6022 \\
 & Question-Emb K-Means & - / - & 0.1098 / 0.2856 & 0.0859 / 0.3656 & 0.0615 / 0.5338 & 0.0522 / \textbf{0.6263} \\
 & k-Center Greedy & - / - & 0.1232 / 0.2197 & 0.0821 / 0.4036 & 0.0608 / 0.5355 & 0.0565 / 0.6151 \\
\rowcolor{orange!15} \cellcolor{white}  & \textbf{CoT-Feature (Ours)} & - / - & \textbf{0.1044} / \textbf{0.4334} & \textbf{0.0731} / \textbf{0.5839} & \textbf{0.0601} / \textbf{0.5646} & \textbf{0.0516} / 0.6013 \\
\cmidrule(lr){2-7}
& \multicolumn{6}{c}{\cellcolor{gray!15}\textit{Trainable Methods (Require Historical Data, N=50)}} \\
\rowcolor{blue!10} \cellcolor{white}  & Correctness K-Means & - / - & 0.1269 / 0.2524 & 0.0942 / 0.3297 & 0.0771 / 0.4398 & 0.0666 / 0.4601 \\
\rowcolor{blue!10} \cellcolor{white}  & IRT & - / - & 0.1047 / 0.3878 & 0.0780 / 0.4071 & 0.0656 / 0.5104 & 0.0540 / 0.5748 \\
\bottomrule
\end{tabular}
}
\end{table}

\begin{table}[H]
\centering
\caption{Performance comparison at N=50 using GPIRT proxy. Best performance among Training-Free methods is highlighted in \textbf{bold}. Trainable methods are shown for reference only.}
\resizebox{\textwidth}{!}{
\begin{tabular}{lcccccc}
\toprule
\multirow{2}{*}{\textbf{Benchmark}} & \multirow{2}{*}{\textbf{Method}} & \multicolumn{5}{c}{\textbf{Sampling Ratio}} \\
\cmidrule(lr){3-7}
& & \makecell{1\% \\ \scriptsize MAE ($\downarrow$) / Sim ($\uparrow$)} & \makecell{5\% \\ \scriptsize MAE ($\downarrow$) / Sim ($\uparrow$)} & \makecell{10\% \\ \scriptsize MAE ($\downarrow$) / Sim ($\uparrow$)} & \makecell{15\% \\ \scriptsize MAE ($\downarrow$) / Sim ($\uparrow$)} & \makecell{20\% \\ \scriptsize MAE ($\downarrow$) / Sim ($\uparrow$)} \\
\midrule
& \multicolumn{6}{c}{\cellcolor{gray!15}\textit{Training-Free Methods (Zero-Shot)}} \\
GSM8K & Random & 0.0933 / \textbf{0.8809} & 0.0325 / 0.9670 & 0.0203 / 0.9839 & \textbf{0.0157} / 0.9894 & 0.0129 / 0.9917 \\
 & Question-Emb K-Means & \textbf{0.0781} / 0.8796 & 0.0310 / \textbf{0.9679} & 0.0204 / 0.9815 & 0.0161 / 0.9871 & \textbf{0.0123} / \textbf{0.9930} \\
 & k-Center Greedy & 0.0972 / 0.8676 & 0.0333 / 0.9658 & 0.0214 / 0.9833 & 0.0167 / \textbf{0.9899} & 0.0143 / 0.9929 \\
\rowcolor{orange!15} \cellcolor{white}  & \textbf{CoT-Feature (Ours)} & 0.0849 / 0.8713 & \textbf{0.0297} / 0.9592 & \textbf{0.0192} / \textbf{0.9856} & 0.0163 / 0.9893 & 0.0133 / 0.9918 \\
\cmidrule(lr){2-7}
& \multicolumn{6}{c}{\cellcolor{gray!15}\textit{Trainable Methods (Require Historical Data, N=50)}} \\
\rowcolor{blue!10} \cellcolor{white}  & Correctness K-Means & 0.0698 / 0.9103 & 0.0292 / 0.9718 & 0.0183 / 0.9855 & 0.0152 / 0.9887 & 0.0140 / 0.9926 \\
\rowcolor{blue!10} \cellcolor{white}  & IRT & 0.0805 / 0.8858 & 0.0278 / 0.9642 & 0.0191 / 0.9844 & 0.0149 / 0.9910 & 0.0128 / 0.9924 \\
\midrule
& \multicolumn{6}{c}{\cellcolor{gray!15}\textit{Training-Free Methods (Zero-Shot)}} \\
MMLU & Random & \textbf{0.0207} / \textbf{0.9658} & 0.0106 / \textbf{0.9908} & 0.0082 / 0.9945 & 0.0067 / 0.9961 & \textbf{0.0056} / 0.9971 \\
 & Question-Emb K-Means & 0.0220 / 0.9597 & 0.0123 / 0.9899 & 0.0108 / 0.9934 & 0.0146 / 0.9948 & 0.0192 / 0.9961 \\
 & k-Center Greedy & 0.0208 / 0.9651 & 0.0239 / 0.9892 & 0.0244 / 0.9925 & 0.0257 / 0.9945 & 0.0244 / 0.9955 \\
\rowcolor{orange!15} \cellcolor{white}  & \textbf{CoT-Feature (Ours)} & 0.0233 / 0.9535 & \textbf{0.0100} / 0.9900 & \textbf{0.0068} / \textbf{0.9949} & \textbf{0.0066} / \textbf{0.9965} & 0.0056 / \textbf{0.9971} \\
\cmidrule(lr){2-7}
& \multicolumn{6}{c}{\cellcolor{gray!15}\textit{Trainable Methods (Require Historical Data, N=50)}} \\
\rowcolor{blue!10} \cellcolor{white}  & Correctness K-Means & 0.0207 / 0.9723 & 0.0141 / 0.9868 & 0.0129 / 0.9887 & 0.0125 / 0.9892 & 0.0121 / 0.9894 \\
\rowcolor{blue!10} \cellcolor{white}  & IRT & 0.0213 / 0.9655 & 0.0105 / 0.9911 & 0.0077 / 0.9949 & 0.0066 / 0.9964 & 0.0065 / 0.9971 \\
\midrule
& \multicolumn{6}{c}{\cellcolor{gray!15}\textit{Training-Free Methods (Zero-Shot)}} \\
MMLU-Pro & Random & 0.0239 / 0.9624 & \textbf{0.0097} / \textbf{0.9906} & 0.0072 / 0.9943 & \textbf{0.0058} / 0.9957 & 0.0049 / 0.9967 \\
 & Question-Emb K-Means & 0.0238 / 0.9593 & 0.0098 / 0.9899 & 0.0075 / \textbf{0.9944} & 0.0073 / 0.9957 & 0.0069 / \textbf{0.9968} \\
 & k-Center Greedy & 0.0249 / 0.9582 & 0.0146 / 0.9895 & 0.0149 / 0.9936 & 0.0163 / 0.9947 & 0.0163 / 0.9956 \\
\rowcolor{orange!15} \cellcolor{white}  & \textbf{CoT-Feature (Ours)} & \textbf{0.0226} / \textbf{0.9691} & 0.0105 / 0.9899 & \textbf{0.0070} / 0.9939 & 0.0059 / \textbf{0.9959} & \textbf{0.0047} / 0.9966 \\
\cmidrule(lr){2-7}
& \multicolumn{6}{c}{\cellcolor{gray!15}\textit{Trainable Methods (Require Historical Data, N=50)}} \\
\rowcolor{blue!10} \cellcolor{white}  & Correctness K-Means & 0.0226 / 0.9668 & 0.0102 / 0.9909 & 0.0103 / 0.9946 & 0.0101 / 0.9951 & 0.0093 / 0.9954 \\
\rowcolor{blue!10} \cellcolor{white}  & IRT & 0.0278 / 0.9518 & 0.0100 / 0.9914 & 0.0083 / 0.9947 & 0.0079 / 0.9958 & 0.0076 / 0.9966 \\
\midrule
& \multicolumn{6}{c}{\cellcolor{gray!15}\textit{Training-Free Methods (Zero-Shot)}} \\
GPQA & Random & - / - & 0.1281 / 0.2834 & \textbf{0.0596} / 0.4492 & 0.0414 / \textbf{0.6098} & \textbf{0.0352} / \textbf{0.6768} \\
 & Question-Emb K-Means & - / - & 0.1370 / 0.2460 & 0.0601 / 0.4870 & \textbf{0.0379} / 0.6004 & 0.0362 / 0.6651 \\
 & k-Center Greedy & - / - & \textbf{0.1172} / 0.3289 & 0.0612 / 0.4532 & 0.0434 / 0.5656 & 0.0364 / 0.6267 \\
\rowcolor{orange!15} \cellcolor{white}  & \textbf{CoT-Feature (Ours)} & - / - & 0.1242 / \textbf{0.3571} & 0.0612 / \textbf{0.5275} & 0.0412 / 0.5904 & 0.0362 / 0.6616 \\
\cmidrule(lr){2-7}
& \multicolumn{6}{c}{\cellcolor{gray!15}\textit{Trainable Methods (Require Historical Data, N=50)}} \\
\rowcolor{blue!10} \cellcolor{white}  & Correctness K-Means & - / - & 0.1028 / 0.3905 & 0.0491 / 0.4433 & 0.0382 / 0.5559 & 0.0373 / 0.6144 \\
\rowcolor{blue!10} \cellcolor{white}  & IRT & - / - & 0.1117 / 0.3240 & 0.0505 / 0.4854 & 0.0411 / 0.6455 & 0.0352 / 0.6663 \\
\bottomrule
\end{tabular}
}
\end{table}

\begin{table}[H]
\centering
\caption{Performance comparison at N=100 using SA proxy. Best performance among Training-Free methods is highlighted in \textbf{bold}. Trainable methods are shown for reference only.}
\resizebox{\textwidth}{!}{
\begin{tabular}{lcccccc}
\toprule
\multirow{2}{*}{\textbf{Benchmark}} & \multirow{2}{*}{\textbf{Method}} & \multicolumn{5}{c}{\textbf{Sampling Ratio}} \\
\cmidrule(lr){3-7}
& & \makecell{1\% \\ \scriptsize MAE ($\downarrow$) / Sim ($\uparrow$)} & \makecell{5\% \\ \scriptsize MAE ($\downarrow$) / Sim ($\uparrow$)} & \makecell{10\% \\ \scriptsize MAE ($\downarrow$) / Sim ($\uparrow$)} & \makecell{15\% \\ \scriptsize MAE ($\downarrow$) / Sim ($\uparrow$)} & \makecell{20\% \\ \scriptsize MAE ($\downarrow$) / Sim ($\uparrow$)} \\
\midrule
& \multicolumn{6}{c}{\cellcolor{gray!15}\textit{Training-Free Methods (Zero-Shot)}} \\
GSM8K & Random & 0.0890 / 0.9044 & 0.0382 / \textbf{0.9747} & 0.0246 / 0.9851 & 0.0187 / \textbf{0.9897} & 0.0142 / 0.9920 \\
 & Question-Emb K-Means & 0.0875 / 0.8951 & \textbf{0.0321} / 0.9743 & 0.0241 / 0.9797 & \textbf{0.0175} / 0.9871 & \textbf{0.0133} / \textbf{0.9931} \\
 & k-Center Greedy & \textbf{0.0683} / \textbf{0.9118} & 0.0349 / 0.9738 & 0.0283 / 0.9847 & 0.0218 / \textbf{0.9897} & 0.0175 / 0.9925 \\
\rowcolor{orange!15} \cellcolor{white}  & \textbf{CoT-Feature (Ours)} & 0.0755 / 0.8884 & 0.0331 / 0.9696 & \textbf{0.0209} / \textbf{0.9853} & 0.0197 / 0.9884 & 0.0154 / 0.9918 \\
\cmidrule(lr){2-7}
& \multicolumn{6}{c}{\cellcolor{gray!15}\textit{Trainable Methods (Require Historical Data, N=100)}} \\
\rowcolor{blue!10} \cellcolor{white}  & Correctness K-Means & 0.0522 / 0.9226 & 0.0344 / 0.9699 & 0.0299 / 0.9843 & 0.0266 / 0.9865 & 0.0233 / 0.9903 \\
\rowcolor{blue!10} \cellcolor{white}  & IRT & 0.0650 / 0.9049 & 0.0295 / 0.9734 & 0.0216 / 0.9797 & 0.0168 / 0.9880 & 0.0141 / 0.9893 \\
\midrule
& \multicolumn{6}{c}{\cellcolor{gray!15}\textit{Training-Free Methods (Zero-Shot)}} \\
MMLU & Random & 0.0330 / \textbf{0.9489} & 0.0149 / 0.9880 & 0.0102 / 0.9935 & \textbf{0.0079} / 0.9957 & 0.0063 / 0.9969 \\
 & Question-Emb K-Means & \textbf{0.0298} / 0.9484 & 0.0215 / 0.9877 & 0.0154 / 0.9923 & 0.0203 / 0.9943 & 0.0252 / 0.9958 \\
 & k-Center Greedy & 0.0477 / 0.9461 & 0.0530 / 0.9867 & 0.0393 / 0.9909 & 0.0360 / 0.9942 & 0.0317 / 0.9952 \\
\rowcolor{orange!15} \cellcolor{white}  & \textbf{CoT-Feature (Ours)} & 0.0324 / 0.9353 & \textbf{0.0118} / \textbf{0.9881} & \textbf{0.0081} / \textbf{0.9941} & 0.0081 / \textbf{0.9964} & \textbf{0.0060} / \textbf{0.9971} \\
\cmidrule(lr){2-7}
& \multicolumn{6}{c}{\cellcolor{gray!15}\textit{Trainable Methods (Require Historical Data, N=100)}} \\
\rowcolor{blue!10} \cellcolor{white}  & Correctness K-Means & 0.0184 / 0.9762 & 0.0111 / 0.9906 & 0.0100 / 0.9917 & 0.0091 / 0.9927 & 0.0082 / 0.9936 \\
\rowcolor{blue!10} \cellcolor{white}  & IRT & 0.0222 / 0.9565 & 0.0105 / 0.9895 & 0.0089 / 0.9932 & 0.0071 / 0.9950 & 0.0063 / 0.9962 \\
\midrule
& \multicolumn{6}{c}{\cellcolor{gray!15}\textit{Training-Free Methods (Zero-Shot)}} \\
MMLU-Pro & Random & 0.0311 / 0.9556 & 0.0132 / \textbf{0.9899} & 0.0093 / 0.9941 & \textbf{0.0070} / 0.9956 & 0.0058 / 0.9967 \\
 & Question-Emb K-Means & 0.0300 / 0.9579 & 0.0133 / 0.9890 & 0.0111 / \textbf{0.9947} & 0.0116 / 0.9957 & 0.0100 / \textbf{0.9969} \\
 & k-Center Greedy & 0.0369 / 0.9524 & 0.0432 / 0.9857 & 0.0334 / 0.9923 & 0.0306 / 0.9936 & 0.0272 / 0.9957 \\
\rowcolor{orange!15} \cellcolor{white}  & \textbf{CoT-Feature (Ours)} & \textbf{0.0296} / \textbf{0.9682} & \textbf{0.0124} / 0.9893 & \textbf{0.0082} / 0.9941 & 0.0072 / \textbf{0.9965} & \textbf{0.0055} / \textbf{0.9969} \\
\cmidrule(lr){2-7}
& \multicolumn{6}{c}{\cellcolor{gray!15}\textit{Trainable Methods (Require Historical Data, N=100)}} \\
\rowcolor{blue!10} \cellcolor{white}  & Correctness K-Means & 0.0289 / 0.9815 & 0.0219 / 0.9919 & 0.0192 / 0.9937 & 0.0175 / 0.9940 & 0.0157 / 0.9948 \\
\rowcolor{blue!10} \cellcolor{white}  & IRT & 0.0274 / 0.9561 & 0.0111 / 0.9923 & 0.0085 / 0.9938 & 0.0064 / 0.9953 & 0.0058 / 0.9961 \\
\midrule
& \multicolumn{6}{c}{\cellcolor{gray!15}\textit{Training-Free Methods (Zero-Shot)}} \\
GPQA & Random & - / - & 0.1252 / 0.2851 & 0.0797 / 0.4480 & 0.0635 / 0.5579 & 0.0536 / 0.6022 \\
 & Question-Emb K-Means & - / - & 0.1098 / 0.2856 & 0.0859 / 0.3656 & 0.0615 / 0.5338 & 0.0522 / \textbf{0.6263} \\
 & k-Center Greedy & - / - & 0.1232 / 0.2197 & 0.0821 / 0.4036 & 0.0608 / 0.5355 & 0.0565 / 0.6151 \\
\rowcolor{orange!15} \cellcolor{white}  & \textbf{CoT-Feature (Ours)} & - / - & \textbf{0.1044} / \textbf{0.4334} & \textbf{0.0731} / \textbf{0.5839} & \textbf{0.0601} / \textbf{0.5646} & \textbf{0.0516} / 0.6013 \\
\cmidrule(lr){2-7}
& \multicolumn{6}{c}{\cellcolor{gray!15}\textit{Trainable Methods (Require Historical Data, N=100)}} \\
\rowcolor{blue!10} \cellcolor{white}  & Correctness K-Means & - / - & 0.1124 / 0.4882 & 0.0925 / 0.4670 & 0.0792 / 0.5425 & 0.0742 / 0.5639 \\
\rowcolor{blue!10} \cellcolor{white}  & IRT & - / - & 0.1125 / 0.3425 & 0.0713 / 0.4955 & 0.0511 / 0.6022 & 0.0466 / 0.6193 \\
\bottomrule
\end{tabular}
}
\end{table}

\begin{table}[H]
\centering
\caption{Performance comparison at N=100 using GPIRT proxy. Best performance among Training-Free methods is highlighted in \textbf{bold}. Trainable methods are shown for reference only.}
\resizebox{\textwidth}{!}{
\begin{tabular}{lcccccc}
\toprule
\multirow{2}{*}{\textbf{Benchmark}} & \multirow{2}{*}{\textbf{Method}} & \multicolumn{5}{c}{\textbf{Sampling Ratio}} \\
\cmidrule(lr){3-7}
& & \makecell{1\% \\ \scriptsize MAE ($\downarrow$) / Sim ($\uparrow$)} & \makecell{5\% \\ \scriptsize MAE ($\downarrow$) / Sim ($\uparrow$)} & \makecell{10\% \\ \scriptsize MAE ($\downarrow$) / Sim ($\uparrow$)} & \makecell{15\% \\ \scriptsize MAE ($\downarrow$) / Sim ($\uparrow$)} & \makecell{20\% \\ \scriptsize MAE ($\downarrow$) / Sim ($\uparrow$)} \\
\midrule
& \multicolumn{6}{c}{\cellcolor{gray!15}\textit{Training-Free Methods (Zero-Shot)}} \\
GSM8K & Random & 0.0881 / \textbf{0.8824} & 0.0292 / \textbf{0.9732} & 0.0200 / 0.9856 & 0.0161 / \textbf{0.9899} & 0.0130 / 0.9921 \\
 & Question-Emb K-Means & 0.0899 / 0.8723 & \textbf{0.0276} / 0.9723 & 0.0210 / 0.9804 & \textbf{0.0158} / 0.9877 & \textbf{0.0125} / \textbf{0.9931} \\
 & k-Center Greedy & 0.0831 / 0.8703 & 0.0279 / 0.9725 & 0.0216 / 0.9832 & 0.0177 / 0.9897 & 0.0149 / 0.9925 \\
\rowcolor{orange!15} \cellcolor{white}  & \textbf{CoT-Feature (Ours)} & \textbf{0.0818} / 0.8751 & 0.0285 / 0.9664 & \textbf{0.0188} / \textbf{0.9860} & 0.0162 / 0.9889 & 0.0135 / 0.9917 \\
\cmidrule(lr){2-7}
& \multicolumn{6}{c}{\cellcolor{gray!15}\textit{Trainable Methods (Require Historical Data, N=100)}} \\
\rowcolor{blue!10} \cellcolor{white}  & Correctness K-Means & 0.0741 / 0.9099 & 0.0263 / 0.9727 & 0.0205 / 0.9893 & 0.0195 / 0.9910 & 0.0177 / 0.9934 \\
\rowcolor{blue!10} \cellcolor{white}  & IRT & 0.0651 / 0.8984 & 0.0260 / 0.9748 & 0.0195 / 0.9809 & 0.0152 / 0.9894 & 0.0128 / 0.9902 \\
\midrule
& \multicolumn{6}{c}{\cellcolor{gray!15}\textit{Training-Free Methods (Zero-Shot)}} \\
MMLU & Random & \textbf{0.0199} / 0.9662 & 0.0098 / 0.9909 & 0.0076 / 0.9944 & 0.0062 / 0.9960 & \textbf{0.0053} / 0.9970 \\
 & Question-Emb K-Means & 0.0206 / 0.9626 & 0.0107 / 0.9895 & 0.0086 / 0.9940 & 0.0103 / 0.9949 & 0.0132 / 0.9962 \\
 & k-Center Greedy & 0.0209 / \textbf{0.9684} & 0.0129 / 0.9902 & 0.0141 / 0.9935 & 0.0161 / 0.9953 & 0.0164 / 0.9959 \\
\rowcolor{orange!15} \cellcolor{white}  & \textbf{CoT-Feature (Ours)} & 0.0218 / 0.9631 & \textbf{0.0096} / \textbf{0.9911} & \textbf{0.0069} / \textbf{0.9952} & \textbf{0.0061} / \textbf{0.9964} & 0.0054 / \textbf{0.9971} \\
\cmidrule(lr){2-7}
& \multicolumn{6}{c}{\cellcolor{gray!15}\textit{Trainable Methods (Require Historical Data, N=100)}} \\
\rowcolor{blue!10} \cellcolor{white}  & Correctness K-Means & 0.0190 / 0.9715 & 0.0100 / 0.9906 & 0.0083 / 0.9932 & 0.0075 / 0.9941 & 0.0066 / 0.9948 \\
\rowcolor{blue!10} \cellcolor{white}  & IRT & 0.0188 / 0.9684 & 0.0099 / 0.9913 & 0.0076 / 0.9945 & 0.0064 / 0.9957 & 0.0057 / 0.9966 \\
\midrule
& \multicolumn{6}{c}{\cellcolor{gray!15}\textit{Training-Free Methods (Zero-Shot)}} \\
MMLU-Pro & Random & 0.0229 / 0.9629 & 0.0094 / \textbf{0.9907} & 0.0066 / \textbf{0.9947} & 0.0053 / 0.9960 & 0.0045 / 0.9969 \\
 & Question-Emb K-Means & 0.0227 / 0.9581 & \textbf{0.0093} / 0.9899 & 0.0069 / 0.9946 & 0.0061 / 0.9959 & 0.0054 / 0.9970 \\
 & k-Center Greedy & 0.0229 / 0.9640 & 0.0102 / 0.9900 & 0.0086 / 0.9941 & 0.0096 / 0.9949 & 0.0099 / 0.9958 \\
\rowcolor{orange!15} \cellcolor{white}  & \textbf{CoT-Feature (Ours)} & \textbf{0.0209} / \textbf{0.9706} & 0.0099 / 0.9903 & \textbf{0.0065} / 0.9944 & \textbf{0.0051} / \textbf{0.9965} & \textbf{0.0043} / \textbf{0.9972} \\
\cmidrule(lr){2-7}
& \multicolumn{6}{c}{\cellcolor{gray!15}\textit{Trainable Methods (Require Historical Data, N=100)}} \\
\rowcolor{blue!10} \cellcolor{white}  & Correctness K-Means & 0.0295 / 0.9550 & 0.0103 / 0.9927 & 0.0091 / 0.9956 & 0.0085 / 0.9970 & 0.0086 / 0.9970 \\
\rowcolor{blue!10} \cellcolor{white}  & IRT & 0.0255 / 0.9520 & 0.0088 / 0.9929 & 0.0066 / 0.9948 & 0.0051 / 0.9962 & 0.0046 / 0.9971 \\
\midrule
& \multicolumn{6}{c}{\cellcolor{gray!15}\textit{Training-Free Methods (Zero-Shot)}} \\
GPQA & Random & - / - & 0.1022 / 0.2447 & 0.0549 / 0.4685 & 0.0415 / 0.5990 & 0.0377 / 0.6439 \\
 & Question-Emb K-Means & - / - & 0.0990 / 0.2167 & 0.0557 / 0.3998 & 0.0412 / 0.5598 & 0.0376 / \textbf{0.6586} \\
 & k-Center Greedy & - / - & 0.1040 / 0.1707 & 0.0565 / 0.3994 & 0.0446 / 0.5275 & 0.0411 / 0.5964 \\
\rowcolor{orange!15} \cellcolor{white}  & \textbf{CoT-Feature (Ours)} & - / - & \textbf{0.0836} / \textbf{0.3995} & \textbf{0.0482} / \textbf{0.6417} & \textbf{0.0380} / \textbf{0.6166} & \textbf{0.0369} / 0.6505 \\
\cmidrule(lr){2-7}
& \multicolumn{6}{c}{\cellcolor{gray!15}\textit{Trainable Methods (Require Historical Data, N=100)}} \\
\rowcolor{blue!10} \cellcolor{white}  & Correctness K-Means & - / - & 0.0653 / 0.4379 & 0.0443 / 0.5324 & 0.0415 / 0.5921 & 0.0431 / 0.6339 \\
\rowcolor{blue!10} \cellcolor{white}  & IRT & - / - & 0.0944 / 0.2877 & 0.0490 / 0.5302 & 0.0360 / 0.6203 & 0.0330 / 0.6608 \\
\bottomrule
\end{tabular}
}
\end{table}

\begin{table}[H]
\centering
\caption{Performance comparison at N=200 using SA proxy. Best performance among Training-Free methods is highlighted in \textbf{bold}. Trainable methods are shown for reference only.}
\resizebox{\textwidth}{!}{
\begin{tabular}{lcccccc}
\toprule
\multirow{2}{*}{\textbf{Benchmark}} & \multirow{2}{*}{\textbf{Method}} & \multicolumn{5}{c}{\textbf{Sampling Ratio}} \\
\cmidrule(lr){3-7}
& & \makecell{1\% \\ \scriptsize MAE ($\downarrow$) / Sim ($\uparrow$)} & \makecell{5\% \\ \scriptsize MAE ($\downarrow$) / Sim ($\uparrow$)} & \makecell{10\% \\ \scriptsize MAE ($\downarrow$) / Sim ($\uparrow$)} & \makecell{15\% \\ \scriptsize MAE ($\downarrow$) / Sim ($\uparrow$)} & \makecell{20\% \\ \scriptsize MAE ($\downarrow$) / Sim ($\uparrow$)} \\
\midrule
& \multicolumn{6}{c}{\cellcolor{gray!15}\textit{Training-Free Methods (Zero-Shot)}} \\
GSM8K & Random & 0.0890 / 0.9044 & 0.0382 / \textbf{0.9747} & 0.0246 / 0.9851 & 0.0187 / \textbf{0.9897} & 0.0142 / 0.9920 \\
 & Question-Emb K-Means & 0.0875 / 0.8951 & \textbf{0.0321} / 0.9743 & 0.0241 / 0.9797 & \textbf{0.0175} / 0.9871 & \textbf{0.0133} / \textbf{0.9931} \\
 & k-Center Greedy & \textbf{0.0683} / \textbf{0.9118} & 0.0349 / 0.9738 & 0.0283 / 0.9847 & 0.0218 / \textbf{0.9897} & 0.0175 / 0.9925 \\
\rowcolor{orange!15} \cellcolor{white}  & \textbf{CoT-Feature (Ours)} & 0.0755 / 0.8884 & 0.0331 / 0.9696 & \textbf{0.0209} / \textbf{0.9853} & 0.0197 / 0.9884 & 0.0154 / 0.9918 \\
\cmidrule(lr){2-7}
& \multicolumn{6}{c}{\cellcolor{gray!15}\textit{Trainable Methods (Require Historical Data, N=200)}} \\
\rowcolor{blue!10} \cellcolor{white}  & Correctness K-Means & 0.0546 / 0.9239 & 0.0348 / 0.9694 & 0.0303 / 0.9800 & 0.0277 / 0.9817 & 0.0259 / 0.9870 \\
\rowcolor{blue!10} \cellcolor{white}  & IRT & 0.0723 / 0.9048 & 0.0283 / 0.9757 & 0.0205 / 0.9841 & 0.0172 / 0.9866 & 0.0148 / 0.9895 \\
\midrule
& \multicolumn{6}{c}{\cellcolor{gray!15}\textit{Training-Free Methods (Zero-Shot)}} \\
MMLU & Random & 0.0330 / \textbf{0.9489} & 0.0149 / 0.9880 & 0.0102 / 0.9935 & \textbf{0.0079} / 0.9957 & 0.0063 / 0.9969 \\
 & Question-Emb K-Means & \textbf{0.0298} / 0.9484 & 0.0215 / 0.9877 & 0.0154 / 0.9923 & 0.0203 / 0.9943 & 0.0252 / 0.9958 \\
 & k-Center Greedy & 0.0477 / 0.9461 & 0.0530 / 0.9867 & 0.0393 / 0.9909 & 0.0360 / 0.9942 & 0.0317 / 0.9952 \\
\rowcolor{orange!15} \cellcolor{white}  & \textbf{CoT-Feature (Ours)} & 0.0324 / 0.9353 & \textbf{0.0118} / \textbf{0.9881} & \textbf{0.0081} / \textbf{0.9941} & 0.0081 / \textbf{0.9964} & \textbf{0.0060} / \textbf{0.9971} \\
\cmidrule(lr){2-7}
& \multicolumn{6}{c}{\cellcolor{gray!15}\textit{Trainable Methods (Require Historical Data, N=200)}} \\
\rowcolor{blue!10} \cellcolor{white}  & Correctness K-Means & 0.0211 / 0.9699 & 0.0139 / 0.9885 & 0.0112 / 0.9898 & 0.0099 / 0.9911 & 0.0085 / 0.9935 \\
\rowcolor{blue!10} \cellcolor{white}  & IRT & 0.0223 / 0.9558 & 0.0097 / 0.9901 & 0.0072 / 0.9928 & 0.0060 / 0.9940 & 0.0055 / 0.9956 \\
\midrule
& \multicolumn{6}{c}{\cellcolor{gray!15}\textit{Training-Free Methods (Zero-Shot)}} \\
MMLU-Pro & Random & 0.0311 / 0.9556 & 0.0132 / \textbf{0.9899} & 0.0093 / 0.9941 & \textbf{0.0070} / 0.9956 & 0.0058 / 0.9967 \\
 & Question-Emb K-Means & 0.0300 / 0.9579 & 0.0133 / 0.9890 & 0.0111 / \textbf{0.9947} & 0.0116 / 0.9957 & 0.0100 / \textbf{0.9969} \\
 & k-Center Greedy & 0.0369 / 0.9524 & 0.0432 / 0.9857 & 0.0334 / 0.9923 & 0.0306 / 0.9936 & 0.0272 / 0.9957 \\
\rowcolor{orange!15} \cellcolor{white}  & \textbf{CoT-Feature (Ours)} & \textbf{0.0296} / \textbf{0.9682} & \textbf{0.0124} / 0.9893 & \textbf{0.0082} / 0.9941 & 0.0072 / \textbf{0.9965} & \textbf{0.0055} / \textbf{0.9969} \\
\cmidrule(lr){2-7}
& \multicolumn{6}{c}{\cellcolor{gray!15}\textit{Trainable Methods (Require Historical Data, N=200)}} \\
\rowcolor{blue!10} \cellcolor{white}  & Correctness K-Means & 0.0332 / 0.9847 & 0.0244 / 0.9918 & 0.0222 / 0.9927 & 0.0194 / 0.9937 & 0.0174 / 0.9940 \\
\rowcolor{blue!10} \cellcolor{white}  & IRT & 0.0233 / 0.9582 & 0.0123 / 0.9901 & 0.0080 / 0.9939 & 0.0061 / 0.9957 & 0.0054 / 0.9963 \\
\midrule
& \multicolumn{6}{c}{\cellcolor{gray!15}\textit{Training-Free Methods (Zero-Shot)}} \\
GPQA & Random & - / - & 0.1252 / 0.2851 & 0.0797 / 0.4480 & 0.0635 / 0.5579 & 0.0536 / 0.6022 \\
 & Question-Emb K-Means & - / - & 0.1098 / 0.2856 & 0.0859 / 0.3656 & 0.0615 / 0.5338 & 0.0522 / \textbf{0.6263} \\
 & k-Center Greedy & - / - & 0.1232 / 0.2197 & 0.0821 / 0.4036 & 0.0608 / 0.5355 & 0.0565 / 0.6151 \\
\rowcolor{orange!15} \cellcolor{white}  & \textbf{CoT-Feature (Ours)} & - / - & \textbf{0.1044} / \textbf{0.4334} & \textbf{0.0731} / \textbf{0.5839} & \textbf{0.0601} / \textbf{0.5646} & \textbf{0.0516} / 0.6013 \\
\cmidrule(lr){2-7}
& \multicolumn{6}{c}{\cellcolor{gray!15}\textit{Trainable Methods (Require Historical Data, N=200)}} \\
\rowcolor{blue!10} \cellcolor{white}  & Correctness K-Means & - / - & 0.1261 / 0.2994 & 0.1072 / 0.3593 & 0.0915 / 0.4311 & 0.0839 / 0.4810 \\
\rowcolor{blue!10} \cellcolor{white}  & IRT & - / - & 0.1055 / 0.2956 & 0.0776 / 0.4353 & 0.0610 / 0.5367 & 0.0529 / 0.5793 \\
\bottomrule
\end{tabular}
}
\end{table}

\begin{table}[H]
\centering
\caption{Performance comparison at N=200 using GPIRT proxy. Best performance among Training-Free methods is highlighted in \textbf{bold}. Trainable methods are shown for reference only.}
\resizebox{\textwidth}{!}{
\begin{tabular}{lcccccc}
\toprule
\multirow{2}{*}{\textbf{Benchmark}} & \multirow{2}{*}{\textbf{Method}} & \multicolumn{5}{c}{\textbf{Sampling Ratio}} \\
\cmidrule(lr){3-7}
& & \makecell{1\% \\ \scriptsize MAE ($\downarrow$) / Sim ($\uparrow$)} & \makecell{5\% \\ \scriptsize MAE ($\downarrow$) / Sim ($\uparrow$)} & \makecell{10\% \\ \scriptsize MAE ($\downarrow$) / Sim ($\uparrow$)} & \makecell{15\% \\ \scriptsize MAE ($\downarrow$) / Sim ($\uparrow$)} & \makecell{20\% \\ \scriptsize MAE ($\downarrow$) / Sim ($\uparrow$)} \\
\midrule
& \multicolumn{6}{c}{\cellcolor{gray!15}\textit{Training-Free Methods (Zero-Shot)}} \\
GSM8K & Random & 0.0881 / 0.8761 & 0.0287 / \textbf{0.9744} & 0.0192 / 0.9859 & \textbf{0.0152} / \textbf{0.9898} & 0.0125 / 0.9921 \\
 & Question-Emb K-Means & 0.0869 / 0.8594 & 0.0279 / 0.9729 & 0.0203 / 0.9811 & 0.0153 / 0.9874 & \textbf{0.0121} / \textbf{0.9931} \\
 & k-Center Greedy & 0.0879 / 0.8655 & \textbf{0.0265} / 0.9731 & 0.0192 / 0.9817 & 0.0158 / 0.9891 & 0.0135 / 0.9923 \\
\rowcolor{orange!15} \cellcolor{white}  & \textbf{CoT-Feature (Ours)} & \textbf{0.0738} / \textbf{0.8793} & 0.0278 / 0.9693 & \textbf{0.0187} / \textbf{0.9860} & 0.0157 / 0.9886 & 0.0131 / 0.9914 \\
\cmidrule(lr){2-7}
& \multicolumn{6}{c}{\cellcolor{gray!15}\textit{Trainable Methods (Require Historical Data, N=200)}} \\
\rowcolor{blue!10} \cellcolor{white}  & Correctness K-Means & 0.0705 / 0.9242 & 0.0260 / 0.9762 & 0.0198 / 0.9875 & 0.0175 / 0.9890 & 0.0163 / 0.9918 \\
\rowcolor{blue!10} \cellcolor{white}  & IRT & 0.0706 / 0.9023 & 0.0252 / 0.9758 & 0.0181 / 0.9857 & 0.0142 / 0.9892 & 0.0130 / 0.9916 \\
\midrule
& \multicolumn{6}{c}{\cellcolor{gray!15}\textit{Training-Free Methods (Zero-Shot)}} \\
MMLU & Random & \textbf{0.0186} / 0.9710 & 0.0093 / \textbf{0.9919} & 0.0074 / 0.9948 & 0.0062 / 0.9963 & 0.0053 / 0.9972 \\
 & Question-Emb K-Means & 0.0201 / 0.9657 & 0.0104 / 0.9906 & 0.0090 / 0.9942 & 0.0112 / 0.9953 & 0.0150 / 0.9964 \\
 & k-Center Greedy & 0.0188 / \textbf{0.9720} & 0.0160 / 0.9914 & 0.0173 / 0.9937 & 0.0192 / 0.9955 & 0.0190 / 0.9959 \\
\rowcolor{orange!15} \cellcolor{white}  & \textbf{CoT-Feature (Ours)} & 0.0216 / 0.9617 & \textbf{0.0085} / \textbf{0.9919} & \textbf{0.0063} / \textbf{0.9956} & \textbf{0.0059} / \textbf{0.9966} & \textbf{0.0052} / \textbf{0.9974} \\
\cmidrule(lr){2-7}
& \multicolumn{6}{c}{\cellcolor{gray!15}\textit{Trainable Methods (Require Historical Data, N=200)}} \\
\rowcolor{blue!10} \cellcolor{white}  & Correctness K-Means & 0.0186 / 0.9722 & 0.0094 / 0.9925 & 0.0080 / 0.9938 & 0.0073 / 0.9944 & 0.0064 / 0.9956 \\
\rowcolor{blue!10} \cellcolor{white}  & IRT & 0.0184 / 0.9714 & 0.0086 / 0.9921 & 0.0067 / 0.9938 & 0.0055 / 0.9954 & 0.0049 / 0.9964 \\
\midrule
& \multicolumn{6}{c}{\cellcolor{gray!15}\textit{Training-Free Methods (Zero-Shot)}} \\
MMLU-Pro & Random & 0.0225 / 0.9646 & 0.0094 / \textbf{0.9908} & 0.0069 / 0.9942 & \textbf{0.0055} / \textbf{0.9955} & 0.0048 / 0.9963 \\
 & Question-Emb K-Means & 0.0228 / 0.9611 & \textbf{0.0094} / 0.9904 & 0.0074 / \textbf{0.9943} & 0.0069 / 0.9952 & 0.0062 / \textbf{0.9967} \\
 & k-Center Greedy & 0.0228 / 0.9673 & 0.0117 / 0.9901 & 0.0111 / 0.9940 & 0.0127 / 0.9949 & 0.0127 / 0.9958 \\
\rowcolor{orange!15} \cellcolor{white}  & \textbf{CoT-Feature (Ours)} & \textbf{0.0216} / \textbf{0.9747} & 0.0098 / 0.9907 & \textbf{0.0069} / 0.9937 & 0.0056 / 0.9954 & \textbf{0.0047} / 0.9962 \\
\cmidrule(lr){2-7}
& \multicolumn{6}{c}{\cellcolor{gray!15}\textit{Trainable Methods (Require Historical Data, N=200)}} \\
\rowcolor{blue!10} \cellcolor{white}  & Correctness K-Means & 0.0234 / 0.9685 & 0.0097 / 0.9933 & 0.0092 / 0.9954 & 0.0087 / 0.9966 & 0.0090 / 0.9968 \\
\rowcolor{blue!10} \cellcolor{white}  & IRT & 0.0221 / 0.9619 & 0.0095 / 0.9908 & 0.0072 / 0.9940 & 0.0052 / 0.9958 & 0.0046 / 0.9963 \\
\midrule
& \multicolumn{6}{c}{\cellcolor{gray!15}\textit{Training-Free Methods (Zero-Shot)}} \\
GPQA & Random & - / - & 0.1108 / 0.2931 & 0.0875 / 0.4409 & 0.0578 / 0.5951 & 0.0423 / \textbf{0.6625} \\
 & Question-Emb K-Means & - / - & 0.1111 / 0.2699 & 0.0826 / 0.4144 & 0.0521 / \textbf{0.6110} & 0.0448 / 0.6466 \\
 & k-Center Greedy & - / - & 0.1170 / 0.2745 & 0.0863 / 0.4452 & 0.0623 / 0.5391 & 0.0487 / 0.5753 \\
\rowcolor{orange!15} \cellcolor{white}  & \textbf{CoT-Feature (Ours)} & - / - & \textbf{0.1031} / \textbf{0.3926} & \textbf{0.0737} / \textbf{0.5764} & \textbf{0.0492} / 0.6092 & \textbf{0.0423} / 0.6502 \\
\cmidrule(lr){2-7}
& \multicolumn{6}{c}{\cellcolor{gray!15}\textit{Trainable Methods (Require Historical Data, N=200)}} \\
\rowcolor{blue!10} \cellcolor{white}  & Correctness K-Means & - / - & 0.0805 / 0.3920 & 0.0605 / 0.4511 & 0.0449 / 0.4943 & 0.0406 / 0.5771 \\
\rowcolor{blue!10} \cellcolor{white}  & IRT & - / - & 0.0837 / 0.3378 & 0.0586 / 0.4497 & 0.0484 / 0.5412 & 0.0374 / 0.6340 \\
\bottomrule
\end{tabular}
}
\end{table}

\end{document}